\pdfoutput=1
\documentclass[journal,twoside,web]{ieeecolor}
\usepackage{tmi}
\usepackage{xcolor}
\usepackage{cite}
\usepackage{amsmath,amssymb,amsfonts}
\usepackage{algorithmic}
\usepackage{graphicx}
\usepackage{textcomp}
\usepackage{hyperref}
\usepackage{multirow}
\usepackage{makecell}
\usepackage{booktabs}

\def\BibTeX{{\rm B\kern-.05em{\sc i\kern-.025em b}\kern-.08em
    T\kern-.1667em\lower.7ex\hbox{E}\kern-.125emX}}
\begin{document}
\bstctlcite{IEEEexample:BSTcontrol}
\title{Robust Polyp Detection and Diagnosis through Compositional Prompt-Guided Diffusion Models}
\author{{Jia Yu, Yan Zhu, Peiyao Fu, Tianyi Chen, Junbo Huang, Quanlin Li, Pinghong Zhou, Zhihua Wang, Fei Wu, Shuo Wang, and Xian Yang}
\thanks{\textcolor{black}{J. Yu, Z. Wang, and F. Wu are with Zhejiang University, Hangzhou, and the Shanghai Institute for Advanced Study of Zhejiang University, Shanghai, China. Y. Zhu, P. Fu, Q. Li, and P. Zhou are with the Endoscopy Center, Zhongshan Hospital, Fudan University, and the Shanghai Collaborative Innovation Center of Endoscopy, Shanghai, China. S. Wang, T. Chen, and J. Huang are with the Digital Medical Research Center, School of Basic Medical Sciences, Fudan University, and the Shanghai Key Laboratory of MICCAI, Shanghai, China. X. Yang is with the Alliance Manchester Business School, The University of Manchester, Manchester, U.K., and the Data Science Institute, Imperial College London, London, U.K.}}%
\thanks{\textcolor{black}{This work was supported in part by the National Key Research and Development Program of China (2024YFF1207500), the National Natural Science Foundation of China (82203193), the Shanghai Municipal Education Commission Project for Promoting Research Paradigm Reform and Empowering Disciplinary Advancement through Artificial Intelligence (SOF101020), and the International Science and Technology Cooperation Program under the 2023 Shanghai Action Plan for Science (23410710400). The computations were performed using the CFFF platform of Fudan University.}}%
\thanks{\textcolor{black}{J. Yu and Y. Zhu are co-first authors. Corresponding authors: Shuo Wang (email: shuowang@fudan.edu.cn) and Xian Yang (email: xian.yang@manchester.ac.uk).}}}

\maketitle

\begin{abstract}
Colorectal cancer (CRC) is a significant global health concern, and early detection through screening plays a critical role in reducing mortality. While deep learning models have shown promise in improving polyp detection, classification, and segmentation, their generalization across diverse clinical environments, particularly with out-of-distribution (OOD) data, remains a challenge. Multi-center datasets like PolypGen have been developed to address these issues, but their collection is costly and time-consuming. Traditional data augmentation techniques provide limited variability, failing to capture the complexity of medical images. Diffusion models  have emerged as a promising solution for generating synthetic polyp images, but the image generation process in current models mainly relies on segmentation masks as the condition, limiting their ability to capture the full clinical context. To overcome these limitations, we propose a Progressive Spectrum Diffusion Model (PSDM) that integrates diverse clinical annotations—such as segmentation masks, bounding boxes, and colonoscopy reports—by transforming them into compositional prompts. These prompts are organized into coarse and fine components, allowing the model to capture both broad spatial structures and fine details, generating clinically accurate synthetic images. By augmenting training data with PSDM-generated samples, our model significantly improves polyp detection, classification, and segmentation. For instance, on the PolypGen dataset, PSDM increases the F1 score by 2.12\% and the mean average precision by 3.09\%, demonstrating superior performance in OOD scenarios and enhanced generalization.
\end{abstract}


\section{Introduction}
\label{sec:introduction}

\IEEEPARstart{C}{olorectal} cancer (CRC) is a major global health concern, ranking as the third most common cancer and the second leading cause of cancer deaths worldwide \cite{bray2018global}. Early screening is vital for reducing mortality by enabling the removal of adenomatous polyps before malignancy \cite{barclay2006colonoscopic}. However, the complexity of colon anatomy and polyp variability pose significant challenges, exposing limitations in human detection skills \cite{kahi2011prevalence}.
Deep learning models have shown great potential in enhancing polyp detection and diagnosis \cite{urban2018deep}, excelling in real-time colonoscopy analysis. However, their performance on out-of-distribution (OOD) data remains problematic, as domain shifts caused by variations in equipment, patient demographics, and procedures hinder robustness. Additionally, these models may amplify annotation bias, leading to missed diagnoses and fairness issues \cite{lin2023improving}. Improving generalization is thus critical for achieving reliable diagnostic accuracy in clinical practice. Recently, colonoscopy records, including images and text reports, have been explored as valuable resources \cite{wang2023knowledge}.

Efforts to improve generalization focus on collecting diverse, well-annotated datasets, which are critical for reliable deep learning performance. For instance, PolypGen \cite{ali2023multi} captures polyp diversity by combining data from six centers. However, creating such datasets is costly, labor-intensive, and constrained by privacy concerns, limiting scalability \cite{wang2021annotation}. Traditional data augmentation techniques, such as affine transformations (e.g., rotation, flipping, cropping) and color modifications, have been widely used to expand datasets and improve generalization. However, these methods offer limited variability and fail to capture the complexity of medical images or generate novel visual features \cite{frid2018gan}.
To address these limitations, synthetic data generation using models like Generative Adversarial Networks (GANs) \cite{goodfellow2020generative} has emerged as a solution, increasing dataset variability and enhancing generalization. In medical imaging, this approach generates diverse and realistic representations, improving model performance across clinical scenarios. However, GANs often struggle with consistent quality and diversity, limiting their practical effectiveness \cite{dhariwal2021diffusion}. Diffusion models (DMs) provide a promising alternative, refining noise into structured data to generate stable and diverse images. DMs address domain generalization challenges and enable targeted applications through conditional prompts \cite{kazerouni2023diffusion}, improving performance in tasks like skin lesion classification \cite{akrout2023diffusion} and colonoscopy analysis \cite{du2023arsdm}\textcolor{black}{\cite{du2024polypsegdiff}}.
\begin{figure*}[htbp]
    \centering
    \includegraphics[width=0.35\linewidth]{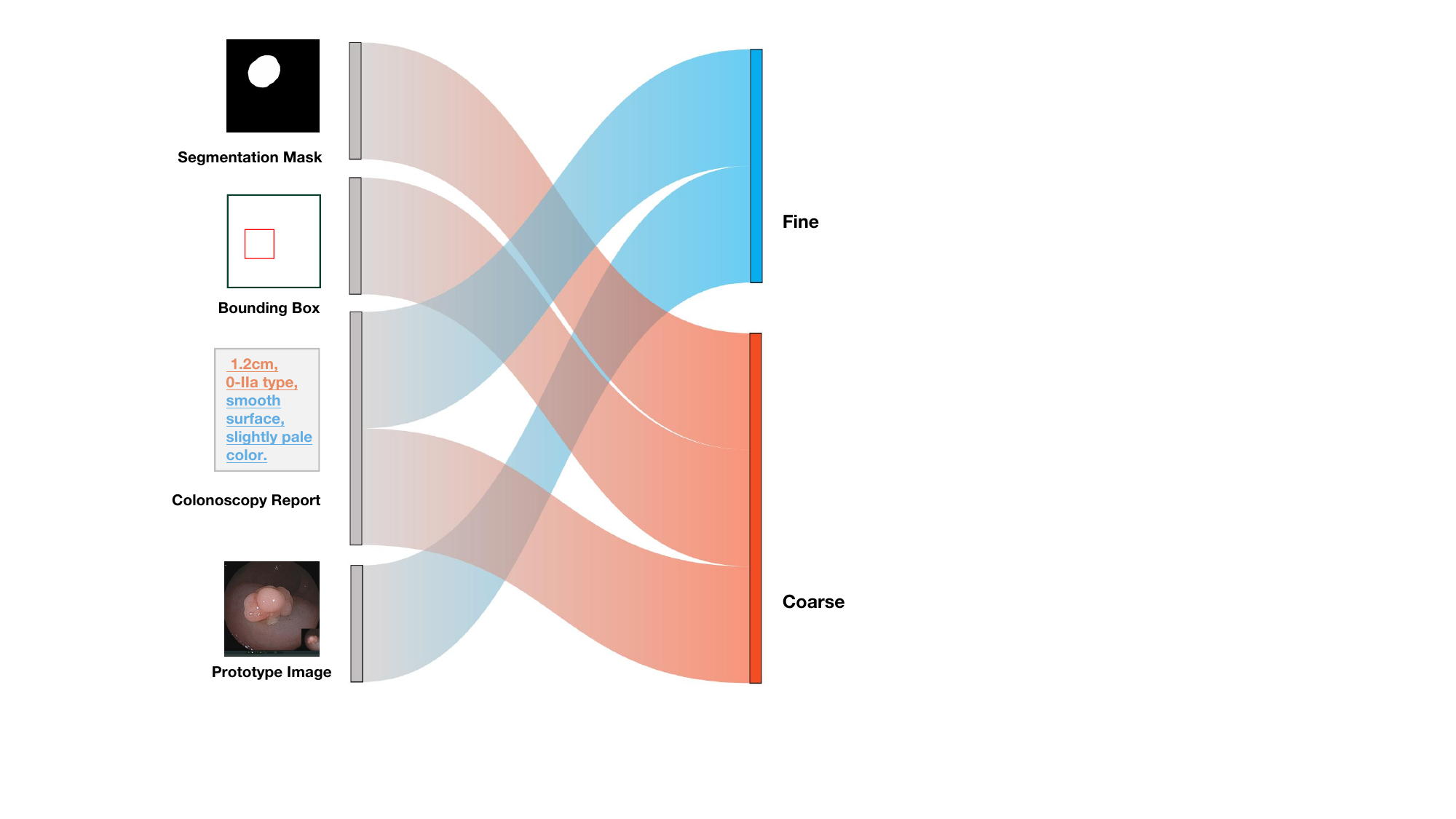}
    \hspace{0.001\linewidth} 
    \includegraphics[width=0.61\linewidth]{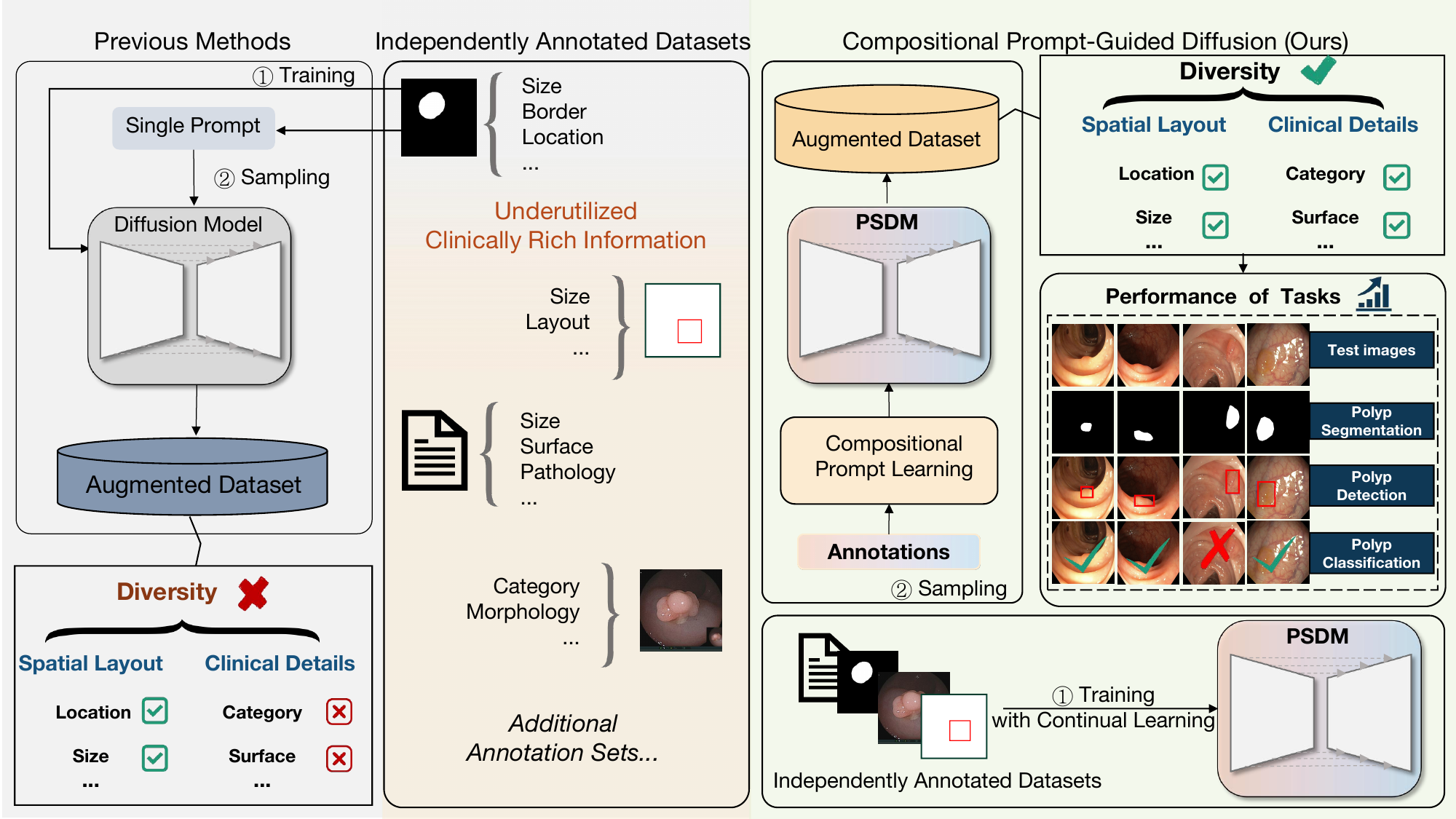}
    \caption{\textcolor{black}{Compositional prompt-guided diffusion framework for generating diverse polyp images. 
    Left: Example prompts with varying levels of granularity. Right: Previous single-prompt methods constrain diversity. In contrast, our PSDM model employs compositional prompts to enhance augmented dataset diversity, resulting in improvements in downstream tasks.}}
    \label{fig:motivation}
\end{figure*}

Despite significant advancements in polyp image generation using diffusion models, a critical limitation remains: these models typically rely heavily on segmentation masks as the primary conditioning input. While segmentation masks provide precise pixel-level information for localization, they lack critical semantic details, such as polyp type, texture, and surrounding tissue characteristics. This reliance limits the models' ability to generate clinically diverse and relevant outputs that reflect real-world polyp variability.
One way to address this limitation is by leveraging additional clinical information, particularly rich textual descriptions from colonoscopy reports and pathology findings, which provide essential semantic insights such as polyp size, morphology, surface characteristics, and histopathological attributes \cite{zhu2023public}. These descriptions, often absent from imaging datasets, offer valuable diagnostic context beyond segmentation masks. Complementary annotations, such as bounding boxes for coarse localization and classification labels for benign or malignant polyps, further enrich the data. However, current models seldom integrate these annotations with segmentation masks, limiting their ability to generate clinically diverse and meaningful outputs \cite{feng2024enhancing}.

As shown in Fig. \ref{fig:motivation}, we propose a compositional prompt approach that integrates diverse annotations, such as segmentation masks, bounding boxes, and colonoscopy reports, into a unified diffusion-based image generation framework. These annotations serve as prompts to guide the diffusion model, enabling the generation of polyp images aligned with the annotations while augmenting data for medical tasks. To effectively integrate complementary information, we organize prompts into coarse components for large-scale spatial structures and fine components for detailed clinical features. To address catastrophic forgetting during sequential training, we adopt continual learning, preserving the model’s ability to generate images from all prompts while learning from new datasets \cite{cao2024controllable}.
Inspired by RQVAE\cite{lee2022autoregressive}, which refines image features sequentially, and diffusion models addressing fine detail degradation during denoising \cite{li2024unveiling}, we adopt a coarse-to-fine image generation process. Our method captures low-frequency structures with coarse components and refines high-frequency details with fine components, ensuring both global layout and intricate details are effectively represented. This hierarchical approach underpins our Progressive Spectrum Diffusion Model (PSDM), which integrates unaligned annotations into a unified framework to generate clinically diverse and relevant colonoscopy images, enhancing generalization and applicability in diverse clinical scenarios.

Thus, our main contributions are as follows:

\begin{itemize} 
    \item \textbf{Colorectal Expert-Annotated Dataset:} In collaboration with colorectal endoscopists, we provide text annotations for open-source datasets, including reports on polyp size, type, surface characteristics, and pathology. \textcolor{black}{All annotations were created by a expert endoscopist, with ambiguous cases further reviewed and reconciled by a second senior clinician to ensure consistency.} This dataset enriches existing public resources for colorectal research.

    \item \textbf{Unaligned Annotation-to-Prompt Integration:} We developed a novel method that integrates diverse clinical annotations into compositional prompts for diffusion models, establishing a framework that harnesses the complementary strengths of various annotations. This approach enhances traditional models by incorporating richer clinical data beyond segmentation masks.

    \item \textbf{Progressive Spectrum Diffusion Model (PSDM):} Our PSDM leverages a frequency-based prompt spectrum, progressing from low-frequency spatial prompts to high-frequency detail prompts. This approach refines polyp images from coarse structures to fine details, generating clinically accurate outputs.

    \item \textbf{Improved Medical Task Performance:} By integrating richer clinical annotations and leveraging PSDM, we significantly enhances performance in polyp classification, detection, and segmentation tasks, improving robustness in handling OOD data for better generalization.
\end{itemize}
\section{Related Work}

\subsection{Generative Adversarial Networks for Polyp Synthesis}

Synthetic polyp generation plays a pivotal role in addressing the scarcity of annotated datasets, a persistent challenge in colonoscopy-related deep learning tasks. Early methods predominantly relied on generative adversarial networks (GANs) due to their ability to synthesize visually realistic images. For example, Shin et al.~\cite{shin2018abnormal} utilized conditional GANs to enhance the realism of synthetic polyps by incorporating edge maps and masks, providing additional structural details to the generated images. Sasmal et al.~\cite{sasmal2020improved} adopted DCGANs to expand polyp datasets, demonstrating improvements in downstream classification performance. Similarly, Qadir et al.~\cite{qadir2022simple} proposed mask-based conditional GANs to manipulate polyp appearances, while He et al.~\cite{he2021colonoscopic} developed adversarial techniques to produce false-negative samples, significantly enhancing the robustness of polyp detection models by challenging classifiers with hard-to-identify cases.

Despite these advances, GAN-based methods face inherent challenges, including convergence instability, limited image diversity, and artifact generation~\cite{dhariwal2021diffusion}. Further studies by Frid-Adar et al.~\cite{frid2018gan} and Yoon et al.~\cite{yoon2022colonoscopic} explored GANs in medical image synthesis, highlighting both their potential and limitations, particularly in colonoscopy.

\subsection{Diffusion Models for Controlled Polyp Generation}

Diffusion models have emerged as robust alternatives to GANs, providing greater stability in training and generating more diverse and realistic synthetic images. Machacek et al.~\cite{machavcek2023mask} introduced a latent diffusion model conditioned on segmentation masks, marking a significant advancement in synthetic polyp generation by focusing on accurate structural details. Du et al.~\cite{du2023arsdm} further developed this concept with ArSDM, incorporating adaptive mechanisms to enhance lesion-specific focus and employing external models to refine alignment accuracy between synthetic polyps and ground truth masks. These refinements led to improved performance in segmentation and detection tasks. Sharma et al.~\cite{Sharma_2024_CVPR} expanded the scope of diffusion-based methods with ControlPolypNet, a framework designed to generate more realistic images by controlling background details and spatial attributes, such as polyp size, shape, and location, resulting in substantial segmentation performance gains.

Despite these advancements, existing frameworks often overlook the wealth of clinical information, focusing on isolated attributes and limiting their ability to achieve comprehensive control over diverse and clinically significant polyp characteristics. This narrow focus restricts their potential to address real-world challenges such as inter-hospital variability and domain shifts.
Our approach builds on these advancements by integrating semantically rich polyp annotations into a unified diffusion-based framework. By enabling joint control across multiple granularity levels, we achieve the modulation of spatial and semantic features during polyp generation, addressing limitations in existing methods and improving model robustness and adaptability for diverse clinical scenarios.

\section{Methods}
\begin{figure*}[htbp]
\centering
\includegraphics[width=1.0\linewidth]{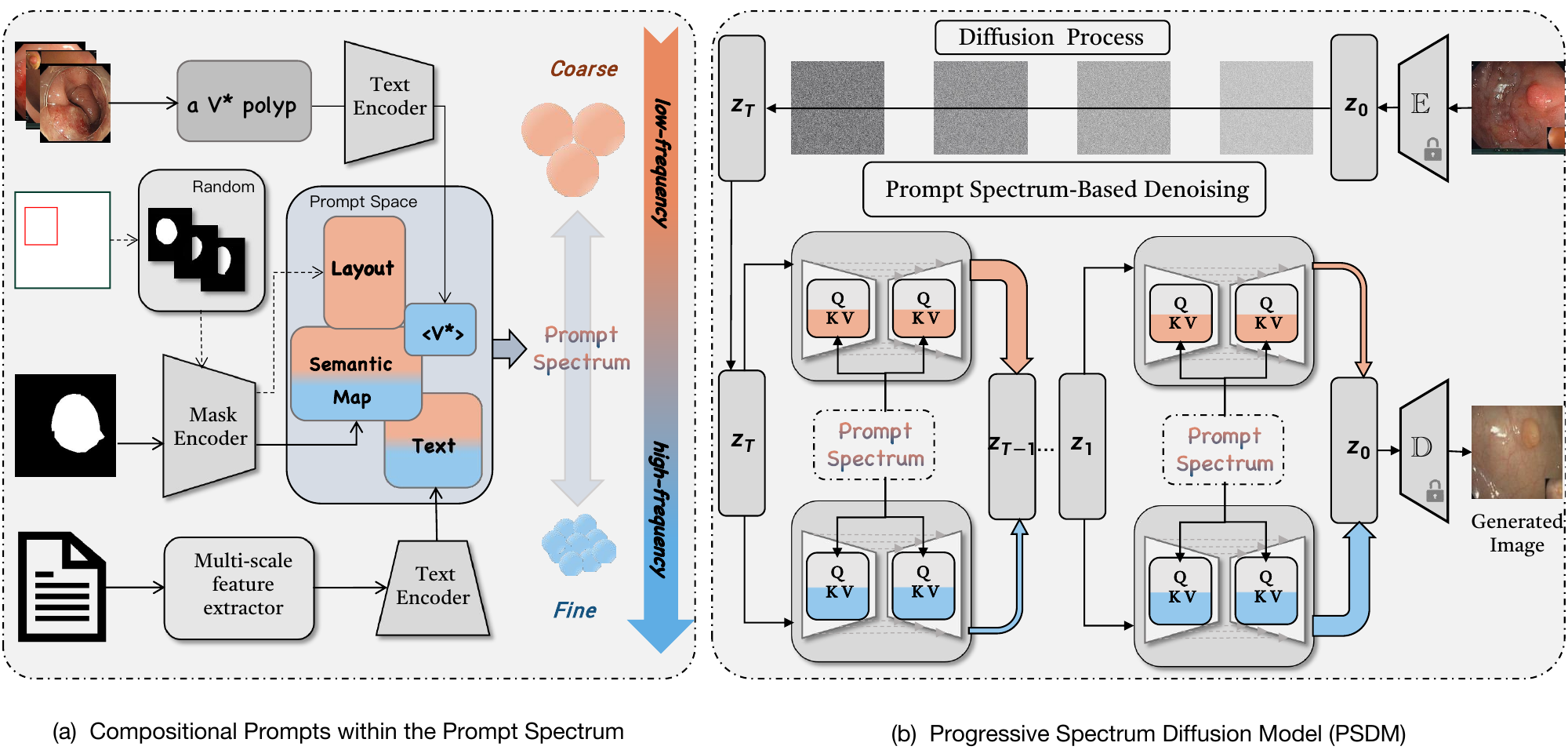}
\caption{\textcolor{black}{The framework integrates diverse annotations into compositional prompts via the \textit{Prompt Spectrum} to capture spatial and clinical details (a, Sec.~\ref{prompts}). During denoising, a conditional U-Net refines latent variables from low to high frequencies, with arrow thickness indicating each component’s contribution (b, Sec.~\ref{diffusion}).}}

\label{framework}
\end{figure*}
\subsection{Model Overview}
\textcolor{black}{Our framework generates high-quality synthetic polyp images by strategically combining compositional prompts derived from diverse annotations, including segmentation masks, bounding boxes, textual descriptions, and prototype images, as illustrated in Fig.~\ref{framework}. Each annotation type is categorized into coarse or fine prompts based on spatial detail and semantic precision, enabling joint control over both localization and morphological realism. Specifically, segmentation masks and bounding boxes constitute coarse prompts, primarily encoding spatial layout and broad localization without intricate appearance details. Textual reports include both coarse spatial characteristics (e.g., polyp size and location) and fine-grained semantic features (e.g., texture, color, polyp type, and pathology). Prototype images specifically address fine-level attributes, especially for rare or uncertain polyp classes with limited detailed annotations.}
To integrate these prompts, we propose the \textit{Prompt Spectrum} (Section~\ref{prompts}), which categorizes the embeddings $\mathbf{P}$ into coarse and fine components to capture both structural layouts and detailed clinical features. These embeddings condition the diffusion model $\mathcal{D}_\theta$ during the \textit{Prompt-guided Diffusion Process} (Section~\ref{diffusion}), enabling a progressive integration of spatial and semantic details to balance global structures and intricate clinical information.

\subsection{Compositional Prompts Generation}\label{prompts}
In this section, we describe the process of generating compositional prompts from various types of medical annotations and how these prompts are integrated into the image generation framework through the \textit{Prompt Spectrum}.

\subsubsection{Prompt Spectrum}
To accommodate the diverse nature of medical annotations, the \textit{Prompt Spectrum} organizes the resulting prompts into two components: coarse and fine. Coarse prompts capture broad, high-level information such as object positions, sizes, and overall layout, while fine prompts focus on detailed aspects, including texture, pathological features, and boundary information.

It is important to note that not all annotations contain a single level of granularity. For annotations that include multi-level information, such as medical reports with coarse-grained size information and fine-grained surface texture details, we process them accordingly (see Section \ref{text}).

By organizing prompts into coarse and fine components, the \textit{Prompt Spectrum} supports the progressive integration of low-frequency structural information and high-frequency clinical details throughout the image generation process. This flexible approach ensures the generation of clinically relevant images that effectively capture the diverse and complementary information encoded in the original annotations.

\subsubsection{Prompt for Multi-Scale Textual Annotations}\label{text}
\begin{figure}[!]
    \centering
    \includegraphics[width=1.0\linewidth]{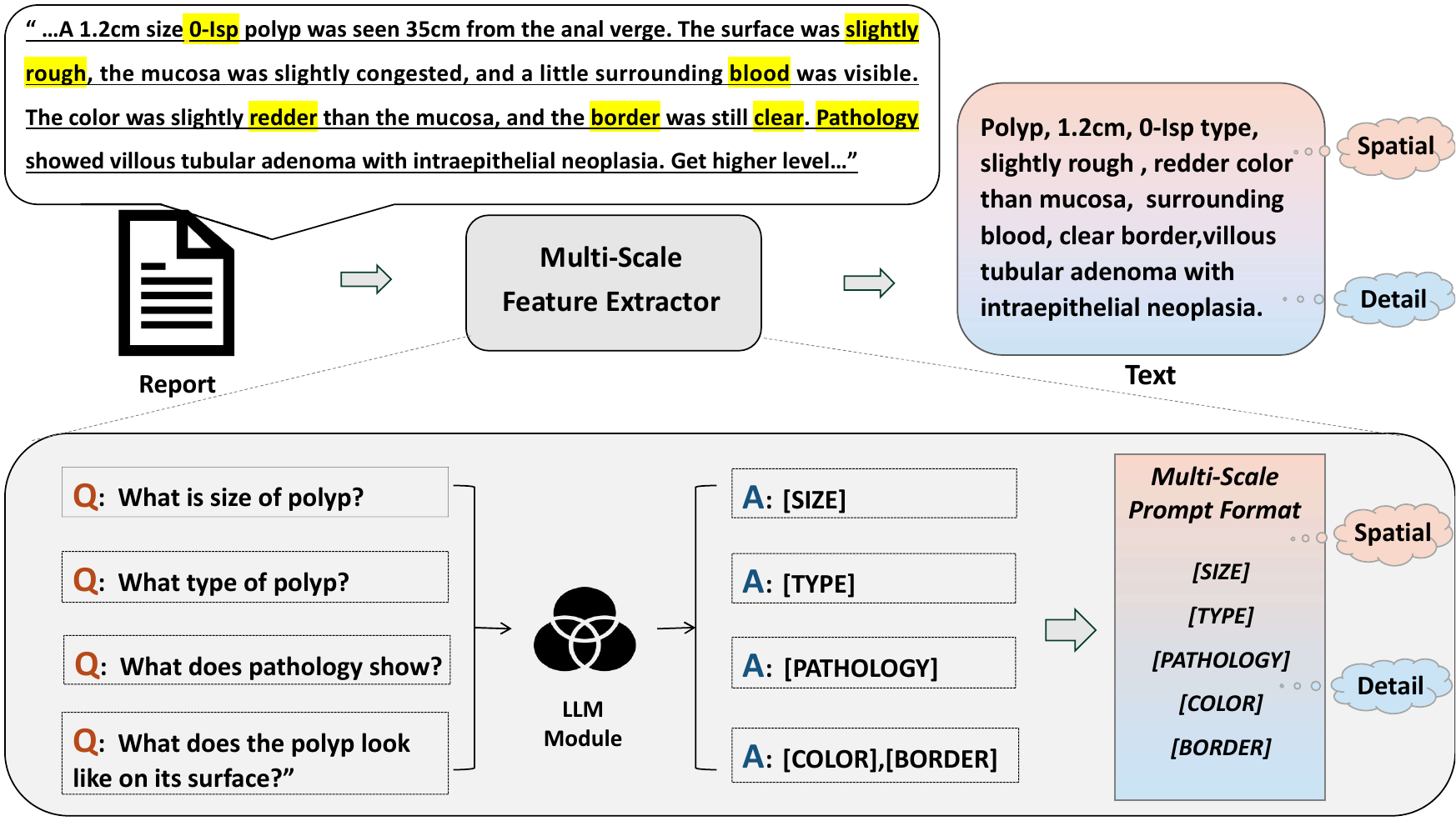}
    \caption{The Multi-Scale Prompt Construction extracts key attributes from medical reports using an LLM, categorizing them into coarse (e.g., size) and fine (e.g., type, pathology, color, border) components.}
\label{fig:LLM}
\end{figure}
Textual annotations, such as clinical reports or descriptive labels, provide crucial contextual information for guiding medical image generation. In this work, we leverage a Large Language Model (LLM) to systematically extract key attributes from these reports and integrate them onto the \textit{Prompt Spectrum}, as illustrated in Fig. \ref{fig:LLM}.

The extraction process involves the LLM responding to a set of predefined, attribute-specific queries targeting essential elements such as size, shape, pathology, color, and location. Coarse spatial attributes, like size and location, are captured by queries such as ``What is the size of the polyp?" or ``Where is the polyp located?" Meanwhile, fine-grained details, such as pathology and surface appearance, are extracted through queries like ``What pathology does the report describe?" or ``What is the surface appearance of the polyp?" 

Once the relevant attributes are extracted, they are incorporated into the \textit{Prompt Spectrum} using structured templates designed to encode both coarse and fine-grained features. Coarse attributes, such as size and location, are formatted into a coarse compositional prompt template, ensuring that the global structure and spatial layout of medical objects are accurately represented in the generated images. For instance, a coarse prompt might follow the format:``[SIZE] polyp located at [LOCATION]", where the extracted size and location are incorporated according to this template.

Similarly, fine-grained attributes—such as pathology, surface texture, and color—are embedded into a fine compositional prompt template. For example, a fine prompt might be formatted as:``polyp with [COLOR] surface and [BORDER] border, diagnosed as [PATHOLOGY]". By organizing the extracted attributes into coarse and fine compositional prompts, the \textit{Prompt Spectrum} enables the model to effectively manage both macro-level structures and intricate clinical details.

We use the CLIP text encoder \cite{radford2021learning}, as configured in Stable Diffusion \cite{rombach2022high}, to convert multi-scale textual attributes into high-dimensional embeddings that guide image generation.

\subsubsection{Prompt for Mask and Bounding Box Annotations}

Encoding segmentation masks and bounding boxes is crucial for capturing the spatial arrangement and structure of regions of interest in generated images. We achieve this through a \texttt{mask\_encoder} that transforms binary masks into compact prompt embeddings for integration into the \textit{Prompt Spectrum}.

The process begins by converting the binary mask to a one-hot representation, distinctly defining foreground and background regions. The one-hot mask is then downsampled through interpolation stages, reducing resolution while preserving key spatial characteristics. A final linear layer produces a compact mask embedding, $\mathbf{p}_{\text{mask}}$.

For bounding boxes, a similar approach is applied. We generate random elliptical masks based on bounding box dimensions, capturing overall spatial structure. 
\textcolor{black}{Specifically, to integrate bounding boxes into our prompt spectrum and support compositional generation without requiring additional encoder training, we approximate coarse spatial layouts by generating an elliptical mask that precisely fits (i.e., tangent to) each bounding box. We then apply slight perturbations along the ellipse boundary to better simulate realistic polyp morphologies observed clinically.}
These masks are processed by the \texttt{mask\_encoder} to produce layout embeddings, $\mathbf{p}_{\text{layout}}$, that represent the bounding box’s size and position without fine-grained detail.

\subsubsection{Prompt Learning for Vague Class Annotations}

In scenarios where class labels are ambiguous or vague, such as when the annotation is not a specific mask or text but rather a \textcolor{black}{prototype} image group representing a vague class, we adopt a \textit{Prompt Learning} approach integrated within the \textit{Prompt Spectrum}. This approach utilizes specialized probabilistic embeddings that allow the model to handle uncertainty by generating prompts that represent a distribution of possible outcomes, rather than a single deterministic label. \textcolor{black}{Our objective is to sample from the distribution defined by these \textcolor{black}{prototype} images.}

When using a set of reference images representing a vague class, the phrase ``a \( V^* polyp\)" is introduced as a specialized token within the prompt embedding \( \mathbf{p}_{V^*} \). This token indicates the vague class and fine-tunes the diffusion model $\mathcal{D}_\theta$, establishing a mapping between the reference images \( \textit{I}_{V^*} \) and the corresponding \( \mathbf{p}_{V^*} \).

Formally, the diffusion model $\mathcal{D}_\theta$ is fine-tuned by minimizing the following loss function:
\begin{equation}
\min_{\theta} \mathcal{L}_{V^*} = \mathbb{E}_{\mathbf{z} \sim \mathcal{E}(\textit{I}_{V^*}), \epsilon \sim \mathcal{N}(0, I), t} \left[ \left\| \epsilon - \epsilon_{\theta}(\mathbf{z}_t, t, \mathbf{p}_{V^*}) \right\|^2 \right]
\end{equation}
where \( \mathbf{z} \sim \mathcal{E}(\textit{I}_{V^*}) \) represents the latent variables encoded from the reference image set \( \textit{I}_{V^*} \), \( \epsilon \) is sampled from a standard normal distribution, and \( t \) is the timestep in the diffusion process. The term \( \mathbf{p}_{V^*} \) is the probabilistic prompt embedding for the vague class \( V^* \). The objective is to optimize the model parameters \( \theta \) to predict the noise \( \epsilon \) given the current latent state \( \mathbf{z}_t \), timestep \( t \), and the prompt embedding \( \mathbf{p}_{V^*} \).

By embedding the vague class annotation into a specialized prompt, the model generates diverse and realistic images that align with the variations present in the reference image group. This method ensures that even with ambiguous or vague annotations, the generated images remain clinically relevant and consistent with the input data.

\subsection{Prompt-guided Diffusion Process}\label{diffusion}

In this work, building on compositional prompts and the \textit{Prompt Spectrum}, we extend the Stable Diffusion framework \cite{rombach2022high}, \textcolor{black}{a powerful latent diffusion model that operates in a compact latent space—dramatically reducing memory and computational demands compared to pixel-space diffusion—} into the Progressive Spectrum Diffusion Model (PSDM). PSDM guides the image generation process through multi-granularity prompts \( \mathbf{p}_i \), ensuring alignment with complex medical annotation requirements.

Diffusion models are probabilistic generative models that iteratively denoise a sample drawn from a Gaussian distribution back to a clean image. Specifically, the input image \( x \) is encoded into a latent representation \( \mathbf{z} = \mathcal{E}(x) \) using a pre-trained encoder \( \mathcal{E} \). The latent code \( \mathbf{z} \) then undergoes a forward diffusion process, progressively corrupted by Gaussian noise:
\begin{equation}
\mathbf{z}_t = \sqrt{\alpha_t} \mathbf{z}_0 + \sqrt{1 - \alpha_t} \epsilon
\end{equation}
where \( \epsilon \sim \mathcal{N}(0, I) \) and \( \alpha_t \) controls the noise schedule at time step \( t \). A time-conditional U-Net model is used to predict and remove the added noise at each step, progressively refining \( \mathbf{z}_t \) back to its clean latent representation \( \mathbf{z}_0 \). The clean latent code is then decoded back to an image using the decoder \( D \).

At each diffusion step, the model predicts the noise component based on the compositional prompt embeddings as:
\begin{equation}
\label{lambda}
    {\epsilon}_{\theta}(\mathbf{z}_t, t, \mathbf{p})  = \sum_{i} \lambda_t^i \cdot {\epsilon}_{\theta}(\mathbf{z}_t, t, \mathbf{p}_i)
\end{equation}
where \( \lambda_t^i \) represents the prompt-specific weighting at each time step \( t \), allowing the contribution of different prompts to vary according to their granularity throughout the diffusion process.

When multiple prompts are provided, the model processes each component in stages, gradually incorporating information according to its frequency.  During the denoising process, the image generation is initially influenced by prompts that encode low-frequency aspects of the image. As the diffusion process advances, prompts with higher-frequency details become more influential. The weighting function \( \lambda_t^i \) dynamically modulates the contributions of these prompts over time, ensuring a smooth transition from low-frequency structural elements to high-frequency clinical details.

After obtaining the mixed noise prediction, we apply \textit{classifier-free guidance} \cite{ho2022classifier} to amplify the effect of the combined prompts by subtracting the unconditional noise prediction \( {\epsilon}_{\theta}(\mathbf{z}_t,t, \emptyset) \) and scaling the difference:
\begin{equation}
\label{DDIM}
\hat{\epsilon}_{\theta}(\mathbf{z}_t ,t, \mathbf{p}) = {\epsilon}_{\theta}(\mathbf{z}_t ,t, \emptyset) + s \cdot \left( {\epsilon}_{\theta}(\mathbf{z}_t ,t, \mathbf{p}) - {\epsilon}_{\theta}(\mathbf{z}_t ,t, \emptyset) \right)
\end{equation}
where \( s \) is the guidance scale. This guidance mechanism amplifies the model's adherence to the provided prompts, ensuring that the generated image more accurately follows the specified conditions.

\subsection{Training Strategy}
\label{train}
Generating accurate images from diverse prompts while learning sequentially from different annotation types introduces a significant challenge known as \textit{catastrophic forgetting}. This occurs when, as new annotations are introduced, the model "forgets" how to generate images based on earlier annotations, diminishing its ability to retain previously learned information \cite{wang2024comprehensive, cao2024controllable}. 
\textcolor{black}{While simultaneous training on all annotation modalities within a fixed dataset may appear intuitive, we observe that different modalities require varying numbers of training steps to converge effectively.}

To address this challenge, we begin by defining the overall loss function used during training:
\begin{equation}
\min_{\theta} \mathcal{L} = \mathbb{E}_{\mathbf{z} \sim \mathcal{E}(x), \epsilon \sim \mathcal{N}(0, I), t} \left[ \left\| \epsilon - \epsilon_{\theta}(\mathbf{z}_t, t, \mathbf{p}) \right\|^2 \right]
\end{equation}
To mitigate the risk of catastrophic forgetting, we adopt a rehearsal-based continual learning strategy through a \textit{prompt replay} mechanism. In this approach, previously encountered prompts, denoted as \( \mathbf{p}_{pre} \), are stored in a rehearsal buffer \( \mathcal{R}(\mathbf{p}) \). At each training step, the model samples from this buffer, allowing it to revisit and optimize on earlier prompts. 

The corresponding loss function for this replay mechanism is formulated as follows:
\begin{equation}
    \resizebox{0.91\hsize}{!}{$
    \underset{\theta}{\min} \mathcal{L}_{\text{Replay}} = \mathbb{E}_{\mathbf{z} \sim \mathcal{E}(x), \epsilon \sim \mathcal{N}(0, I), t, \mathbf{p}_{pre} \sim \mathcal{R}(\mathbf{p})} 
    \left[ \left\| \epsilon - \epsilon_{\theta}(\mathbf{z}_t, t, \mathbf{p}_{pre}) \right\|^2 \right]
    $}
\end{equation}

To further enhance the retention of prior knowledge, we implement validation checkpoints throughout the training process. These checkpoints systematically assess the model’s ability to generate images from previously learned prompts, ensuring that the introduction of new annotations does not degrade performance on earlier tasks. This continual learning approach strikes a balance between acquiring new knowledge and retaining prior information, enabling the model to adapt effectively to complex and evolving datasets.

\subsection{Downstream Application}
\label{apply}
To enhance the performance of downstream tasks, it has become common practice to combine synthetic and real data during model training. However, recent studies \cite{wang2024generated} have highlighted the potential risks associated with this approach, noting that an inappropriate mixing ratio between synthetic and real data can lead to performance degradation. This finding underscores the importance of carefully balancing the contribution of each data source during training.

In addition, research has shown that pretraining on in-domain data can significantly improve model performance \cite{ramanujan2024connection, liu2022improved, boers2024foundation}. Building on these insights, we leverage the synthetic data generated by our PSDM during the pretraining phase. 

To generate high-quality synthetic data, we employ Denoising Diffusion Implicit Models (DDIM) \cite{song2020denoising} as the sampler.  Specifically, the DDIM sampling process is governed by the following equation:
\begin{equation}
\begin{aligned}
\mathbf{z}_{t-1} = &\sqrt{\bar{\alpha}_{t-1}} \left( \frac{\mathbf{z}_t  - \sqrt{1 - \bar{\alpha}_t} \hat{\epsilon}_{\theta}(\mathbf{z}_t ,t, \mathbf{p})}{\sqrt{\bar{\alpha}_t}} \right)  \\+ & \sqrt{1 - \bar{\alpha}_{t-1} - \sigma^2} \hat{\epsilon}_{\theta}(\mathbf{z}_t ,t, \mathbf{p})  + \sigma \epsilon
\end{aligned}
\end{equation}
where \( \sigma \) is a noise term. \( \hat{\epsilon}_{\theta}(\mathbf{z}_t ,t, \mathbf{p}) \) is obtained from the model's noise prediction at each timestep \( t \), calculated according to \eqref{DDIM}. This equation ensures that the noisy latent variable \( \mathbf{z}_t \) is progressively refined into \( \mathbf{z}_{t-1} \). Once the refined noise prediction is obtained, the model updates the latent representation \( \mathbf{z}_t \), progressively denoising it until the clean latent code \( \mathbf{z}_0 \) is recovered. The final image is then generated by decoding \( \mathbf{z}_0 \) through the decoder \( D \).

Our PSDM framework flexibly accommodates both single and multiple prompts as input. For segmentation tasks, a single segmentation mask can be transformed into a semantic map, facilitating the generation of synthetic images to enhance pretraining for lesion recognition. Additionally, combining segmentation masks with textual descriptions as compositional prompts further expands the dataset, providing enriched training samples. Following pretraining with synthetic data, fine-tuning on real-world datasets ensures better alignment with true data distributions, thereby improving generalization and enhancing model performance in clinical applications.


\section{Experiments and Results}
\subsection{Experimental Setup}

\subsubsection{Datasets}

\textit{Public Datasets:}
Our experiments utilized a variety of publicly available polyp datasets commonly used for segmentation and detection tasks in medical imaging. These datasets include ETIS \cite{silva2014toward}, CVC-ClinicDB (CVC-612) \cite{bernal2015wm}, CVC-ColonDB \cite{tajbakhsh2015automated}, CVC-300 \cite{vazquez2017benchmark}, Kvasir \cite{jha2020kvasir}, and PolypGen \cite{ali2023multi}.  The PolypGen dataset, which is sourced from six different centers (C1 to C6), offers a broad range of polyp images with corresponding segmentation masks, as well as bounding box annotations for sequential data (Seq1-23).

\textit{Own Datasets:} In addition to the publicly available datasets, we also contributed two datasets specifically designed for our experiments. The first, named the Polyplus dataset, was created by carefully selecting 100 representative polyp images from a larger collection of 1,450 image-mask pairs sourced from Kvasir and CVC-ClinicDB. These images were annotated with detailed textual descriptions by expert colonoscopists, making Polyplus a valuable resource for integrating visual and textual modalities. The second dataset consists of 385 benign and 77 malignant polyp images, annotated with classification labels to support experiments in polyp benign/malignant classification.

The details of these datasets are summarized in Table~\ref{tab:datasets_summary}. The code and datasets associated with our experiments can be accessed via the following link:\href{https://github.com/Jia7878/compositional-prompt-diffusion-for-polyp-generation}{https://github.com/Compositional Prompt-Guided Diffusion}.

\begin{table}[ht]
    \centering
    \caption{Summary of Public and Private Datasets}
    \resizebox{0.50\textwidth}{!}{
    \begin{tabular}{|l|c|c|c|}
        \hline
        \textbf{Dataset} & \textbf{Source} & \textbf{Sample Count} & \textbf{Annotation Type} \\
        \hline
        ETIS & Public & 196 & Segmentation Mask \\
        CVC-ClinicDB (CVC-612) & Public & 612 & Segmentation Mask \\
        CVC-ColonDB & Public & 380 & Segmentation Mask \\
        CVC-300 & Public & 60 & Segmentation Mask \\
        Kvasir & Public & 1,000 & Segmentation Mask \\
        PolypGen (Single frame) & Public & 1,537  & Segmentation Mask \\
        PolypGen (Sequence) & Public & Seq 1-23 & Bounding Box \\
        \hline
        Polyplus & Private & 100 & \makecell[c]{Text Description \\ + Segmentation Mask} \\
        Benign/Malignant Polyp Dataset & Private & 462 & Classification Label \\
        \hline
    \end{tabular}
    }
    \label{tab:datasets_summary}
\end{table}

\subsubsection{Downstream Tasks}
We conducted our experiments across several downstream tasks, including polyp segmentation, polyp benign/malignant classification, and polyp detection. Each task was carefully designed with specific dataset configurations and model settings, as detailed below:

\textit{Polyp Segmentation:}
For the segmentation task, following the PraNet standard, we trained our model on 1,450 image-mask pairs sourced from the Kvasir and CVC-ClinicDB datasets. Additionally, the PolypGen dataset (centers C1–C6), \textcolor{black}{which serves as out-of-distribution data}, was incorporated into the evaluation process to ensure a comprehensive assessment across diverse clinical environments.
To improve the model's generalization, we introduced two augmented datasets: the \textit{Segmentation Mask Only} set, containing 1,450 image-mask pairs generated solely from the original training masks, and the \textit{Segmentation Mask-Text Description} set, with 1,450 pairs generated using both textual descriptions and segmentation masks from the same training set.

\textit{Polyp Classification:}
For the benign/malignant classification task, we used a dataset of 385 benign and 77 malignant polyp images, split into training (308 benign, 61 malignant), validation (38 benign, 7 malignant), and test sets (39 benign, 9 malignant) for model evaluation. To address class imbalance, we augmented the malignant class with an additional 247 images generated by our PSDM.

\textit{Polyp Detection:}
\textcolor{black}{For detection, we used the PolypGen dataset(excluding 64 unlabeled C3 images). Training comprised 5,698 frames (1,385 C1–C5 images, 1,793 positive and 2,520 negative SequenceData frames from Seq1–15 and Seq1–13); testing comprised 2,275 frames (88 C6 images, 432 positive and 1,755 negative frames from Seq16–23 and Seq14–23).}
We also introduced two augmented datasets for detection: the \textit{Bounding Box Only} set, with bounding boxes regressed from the \textit{Segmentation Mask Only} set, and the \textit{Bounding Box-Text Description} set, with bounding boxes from the \textit{Segmentation Mask-Text Description} set. Bounding boxes were defined by the smallest rectangle enclosing each polyp in the segmentation masks.

\subsubsection{Baseline Models and Implementations}

For the segmentation task, we employ three state-of-the-art models: PraNet \cite{fan2020pranet}, Polyp-PVT \cite{dong2021polyp}, and Polyp-CASCADE \cite{rahman2023medical}. PraNet was selected for its proven effectiveness in capturing both local and global features, Polyp-PVT for its superior performance in handling multi-scale features, and Polyp-CASCADE for its robust architecture in medical segmentation tasks.
For the classification task, we adopt ResNet \cite{he2016deep}, a widely used model in medical image analysis known for its effectiveness \textcolor{black}{and better suited than large-parameter Transformers such as ViT for small-sample experiments\cite{lu2022bridging}.}
For the polyp detection task, we utilize YOLOv5 \cite{redmon2016you}, chosen for its high performance and efficiency in real-time object detection, which is critical in clinical settings.
The hyperparameters for each model are provided in Table \ref{tab:hyperparameters}, categorized by task.
\textcolor{black}{Each model was trained for 100 epochs, and the optimal values of the evaluation metrics were documented. Experiments for PSDM used Stable Diffusion 1.5~\cite{rombach2022high} with default settings: a DDIM sampler ($T=200$, guidance scale $w=7.5$) and the frozen CLIP encoder. The U-Net backbone consists of four down- and up-sampling stages, each containing two $3\times3$ residual blocks, with a base channel width of 320 scaled by multipliers [1,2,4,4], and cross-attention layers at the middle two scales, totaling approximately 860 million parameters. All input images were resized to $256\times256$px (mapped to a $64\times64$ latent grid). Experiments ran on a single NVIDIA GeForce RTX 4090 (24 GB VRAM) under Ubuntu 20.04; fine-tuning required about 4 h of wall-clock time, with peak GPU memory usage around 24 GB. Our implementation uses Python 3.8.5 and CUDA 11.7.}

\subsubsection{Comparison with Other Generative Models}
\textcolor{black}{To comprehensively assess the performance of our proposed PSDM, we conduct comparisons with three representative generative models: the GAN-based Pix2Pix~\cite{isola2017image}, the DDPM-based ArSDM~\cite{du2023arsdm}, and the LDM-based Mask\_C~\cite{machavcek2023mask}. For fair and rigorous evaluation, all methods were trained under identical conditions, including the same dataset, random seeds, and classifier-guidance parameters, with an equal number of samples generated for quantitative analysis.}

\begin{table}[ht]
    \centering
    \caption{Hyperparameter Configurations and Evaluation Metrics}
    \resizebox{0.50\textwidth}{!}{
    \begin{tabular}{|l|c|l|l|l|}
        \hline
        \textbf{Task}              & \textbf{Model}        & \textbf{Hyperparameter}       & \textbf{Value}                   & \textbf{Evaluation Metric}          \\ \hline
        \multirow{6}{*}{\centering Segmentation} 
                                   & \multirow{6}{*}{\centering \begin{tabular}[c]{@{}c@{}}PraNet\\ Polyp-PVT\\ Polyp-CASCADE\end{tabular}} 
                                                           & Epoch                         & 100                              & \multirow{6}{*}{mDice, mIoU}         \\ \cline{3-4} 
                                   &                        & Learning Rate                 & 1e-4                             &                                      \\ \cline{3-4} 
                                   &                        & Batch Size                    & 16                               &                                      \\ \cline{3-4} 
                                   &                        & Train Size                    & 352                              &                                      \\ \cline{3-4} 
                                   &                        & LR Decay Epoch                & 50                               &                                      \\ \cline{3-4} 
                                   &                        & Optimizer                     & Adam                            &                                      \\ \hline
        \multirow{4}{*}{\centering Classification} 
                                   & \multirow{4}{*}{\centering ResNet} 
                                                           & Learning Rate                 & 0.001                             & \multirow{4}{*}{\begin{tabular}[c]{@{}l@{}}Precision, Recall, \\ F1-Score, Accuracy, \\ AUC, mAP\end{tabular}} \\ \cline{3-4}
                                   &                        & Batch Size                    & 32                               &                                      \\ \cline{3-4} 
                                   &                        & Epoch                         & 50                               &                                      \\ \cline{3-4} 
                                   &                        & Optimizer                     & Adam                             &                                      \\ \hline
        \multirow{4}{*}{\centering Detection} 
                                   & \multirow{4}{*}{\centering YOLOv5} 
                                                           & Initial Weights               & yolov5s                       & \multirow{4}{*}{\begin{tabular}[c]{@{}l@{}}F1-Score, mAP50, \\ mAP50-95\end{tabular}} \\ \cline{3-4}
                                   &                        & Dataset Format                & coco128                     &                                      \\ \cline{3-4} 
                                   &                        & Batch Size                    & 96                               &                                      \\ \cline{3-4} 
                                   &                        & Epoch                         & 300                              &                                      \\ \hline
    \end{tabular}
    }
    \label{tab:hyperparameters}
\end{table}

\begin{table*}[ht]
\centering
\caption{mDice and mIoU results for standard segmentation datasets. Best results are \textbf{bold}.}
\label{five}
\resizebox{\textwidth}{!}{
\begin{tabular}{c c cc cc cc cc cc cc}
\toprule
\textbf{Model} & \textbf{Method}
  & \multicolumn{2}{c}{\textbf{CVC-300}}
  & \multicolumn{2}{c}{\textbf{Clinic-DB}}
  & \multicolumn{2}{c}{\textbf{Kvasir}}
  & \multicolumn{2}{c}{\textbf{CVC-ColonDB}}
  & \multicolumn{2}{c}{\textbf{ETIS}}
  & \textbf{mDice} & \textbf{mIoU} \\
 & & \textbf{mDice} & \textbf{mIoU}
   & \textbf{mDice} & \textbf{mIoU}
   & \textbf{mDice} & \textbf{mIoU}
   & \textbf{mDice} & \textbf{mIoU}
   & \textbf{mDice} & \textbf{mIoU}
   & \textbf{Overall} & \textbf{Overall} \\
\midrule
\multirow{6}{*}{PraNet}
 & Baseline
   & 88.22 & 81.34
   & 90.83 & 86.23
   & \textbf{90.72} & \textbf{85.10}
   & 71.24 & 64.37
   & 62.25 & 56.15
   & 74.27 & 67.92 \\
 & \textcolor{black}{Pixel to Pixel}
   & \textcolor{black}{\textbf{88.83}} & \textcolor{black}{81.57*}
   & \textcolor{black}{92.09*} & \textcolor{black}{86.67*}
   & \textcolor{black}{88.82*} & \textcolor{black}{83.04}
   & \textcolor{black}{73.67} & \textcolor{black}{65.74}
   & \textcolor{black}{62.84} & \textcolor{black}{56.37}
   & \textcolor{black}{75.47} & \textcolor{black}{68.42} \\
 & \textcolor{black}{ArSDM}
   & \textcolor{black}{86.13*} & \textcolor{black}{79.40*}
   & \textcolor{black}{90.66*} & \textcolor{black}{85.38*}
   & \textcolor{black}{87.92*} & \textcolor{black}{82.29*}
   & \textcolor{black}{72.71*} & \textcolor{black}{65.02*}
   & \textbf{\textcolor{black}{68.91}}\textcolor{black}{*}
     & \textbf{\textcolor{black}{61.66}}\textcolor{black}{*}

   & \textcolor{black}{76.08} & \textcolor{black}{69.02} \\
 & \textcolor{black}{Mask\_C}
   & \textcolor{black}{88.81} & \textcolor{black}{81.58}
   & \textcolor{black}{90.91*} & \textcolor{black}{86.00*}
   & \textcolor{black}{89.86*} & \textcolor{black}{84.37*}
   & \textcolor{black}{72.24} & \textcolor{black}{65.01}
   & \textcolor{black}{57.38} & \textcolor{black}{57.38*}
   & \textcolor{black}{75.31} & \textcolor{black}{68.43} \\
   & $\text{PSDM}^{\text{S}}$
   & 88.37 & 81.31
   & \textbf{92.30}\textcolor{black}{*} & \textbf{87.60}\textcolor{black}{*}
   & 89.76\textcolor{black}{*} & 84.12
   & \textbf{74.00}\textcolor{black}{*} & \textbf{66.45}\textcolor{black}{*}
   & 63.55\textcolor{black}{*} & 56.95\textcolor{black}{*}
   & 75.91 & 69.09 \\

   & $\text{PSDM}^{\text{M}}$
   & 88.49\textcolor{black}{*} & \textbf{81.64}\textcolor{black}{*}
   & 92.08 & 86.99\textcolor{black}{*}
   & 89.37 & 83.72
   & 73.57\textcolor{black}{*} & 66.21\textcolor{black}{*}
   & 67.81\textcolor{black}{*} & 60.98\textcolor{black}{*}
   & \textbf{76.70} & \textbf{69.89} \\
\midrule
\multirow{6}{*}{Polyp-PVT}
 & Baseline
   & 88.80 & 81.65
   & 92.16 & 87.03
   & 90.23 & 85.41
   & 81.28 & 73.05
   & 77.30 & 68.74
   & 82.83 & 75.27 \\
 & \textcolor{black}{Pixel to Pixel}
   & \textcolor{black}{87.45} & \textcolor{black}{79.49}
   & \textcolor{black}{91.71*} & \textcolor{black}{85.87*}
   & \textcolor{black}{91.01} & \textcolor{black}{85.40*}
   & \textcolor{black}{79.82*} & \textcolor{black}{71.28*}
   & \textcolor{black}{76.57*} & \textcolor{black}{67.52*}
   & \textcolor{black}{81.92} & \textcolor{black}{73.88} \\
 & \textcolor{black}{ArSDM}
   & \textcolor{black}{88.18*} & \textcolor{black}{80.54*}
   & \textcolor{black}{92.59*} & \textcolor{black}{87.51*}
   & \textcolor{black}{91.46*} & \textcolor{black}{86.47*}
   & \textcolor{black}{\textbf{81.58}} & \textcolor{black}{73.05}
   & \textcolor{black}{78.24*} & \textcolor{black}{70.08*}
   & \textcolor{black}{83.34} & \textcolor{black}{75.69} \\
 & \textcolor{black}{Mask\_C}
   & \textcolor{black}{88.66*} & \textcolor{black}{81.26*}
   & \textcolor{black}{92.11*} & \textcolor{black}{86.57*}
   & \textcolor{black}{91.66} & \textcolor{black}{86.47}
   & \textcolor{black}{80.57} & \textcolor{black}{71.66*}
   & \textcolor{black}{77.12} & \textcolor{black}{68.12*}
   & \textcolor{black}{82.62} & \textcolor{black}{74.53} \\
   & $\text{PSDM}^{\text{S}}$
   & \textbf{90.74}\textcolor{black}{*} & \textbf{84.33}\textcolor{black}{*}
   & 93.03\textcolor{black}{*} & 88.14\textcolor{black}{*}
   & \textbf{92.09}\textcolor{black}{*} & \textbf{87.40}\textcolor{black}{*}
   & 80.38 & 72.50
   & \textbf{81.22}\textcolor{black}{*} & \textbf{73.13}\textcolor{black}{*}
   & \textbf{83.82} & \textbf{76.63} \\

   & $\text{PSDM}^{\text{M}}$
   & 89.13 & 82.16
   & \textbf{93.28}\textcolor{black}{*} & \textbf{88.39}\textcolor{black}{*}
   & 91.21\textcolor{black}{*} & 86.12
   & 81.23 & \textbf{73.06}
   & 78.73\textcolor{black}{*} & 70.58\textcolor{black}{*}
   & 83.40 & 75.96 \\
\midrule
\multirow{6}{*}{Polyp-CASCADE}
 & Baseline
   & 88.61 & 81.50
   & 93.51 & 88.64
   & \textbf{93.09} & \textbf{88.48}
   & 80.54 & 72.75
   & 78.42 & 70.40
   & 83.21 & 76.04 \\
 & \textcolor{black}{Pixel to Pixel}
   & \textbf{\textcolor{black}{90.34}}\textcolor{black}{*} & \textbf{\textcolor{black}{83.54}}\textcolor{black}{*}

   & \textcolor{black}{93.72*} & \textcolor{black}{89.12*}
   & \textcolor{black}{92.02} & \textcolor{black}{87.13*}
   & \textcolor{black}{81.50} & \textcolor{black}{73.34}
   & \textcolor{black}{79.53*} & \textcolor{black}{71.64*}
   & \textcolor{black}{83.94} & \textcolor{black}{76.64} \\
 & \textcolor{black}{ArSDM}
   & \textcolor{black}{88.99*} & \textcolor{black}{82.11*}
   & \textcolor{black}{93.13*} & \textcolor{black}{88.51*}
   & \textcolor{black}{90.64*} & \textcolor{black}{85.31*}
      & \textbf{\textcolor{black}{82.29}}\textcolor{black}{*} & \textbf{\textcolor{black}{74.20}}\textcolor{black}{*}

   & \textcolor{black}{79.94*} & \textcolor{black}{72.22*}
   & \textcolor{black}{84.10} & \textcolor{black}{76.81} \\
 & \textcolor{black}{Mask\_C}
   & \textcolor{black}{89.49*} & \textcolor{black}{82.79*}
   & \textcolor{black}{93.78*} & \textcolor{black}{89.19*}
   & \textcolor{black}{92.11} & \textcolor{black}{87.11}
   & \textcolor{black}{82.00*} & \textcolor{black}{73.89*}
   & \textcolor{black}{77.89*} & \textcolor{black}{70.26}
   & \textcolor{black}{83.73} & \textcolor{black}{76.51} \\
   & $\text{PSDM}^{\text{S}}$
   & 88.74\textcolor{black}{*} & 82.01\textcolor{black}{*}
   & 93.50 & 88.92
   & 92.34 & 87.66
   & 80.86 & 72.88\textcolor{black}{*}
   & 79.33\textcolor{black}{*} & 71.45\textcolor{black}{*}
   & 83.50 & 76.31 \\

   & $\text{PSDM}^{\text{M}}$
   & 89.26\textcolor{black}{*} & 82.64\textcolor{black}{*}
   & \textbf{94.02}\textcolor{black}{*} & \textbf{89.44}\textcolor{black}{*}
   & 92.01\textcolor{black}{*} & 87.16\textcolor{black}{*}
   & 82.28 & 73.93
   & \textbf{80.31}\textcolor{black}{*} & \textbf{72.51}\textcolor{black}{*}
   & \textbf{84.45} & \textbf{77.10} \\
\bottomrule
\end{tabular}
}
\end{table*}

\begin{table*}[t]
\centering
\caption{mDice and mIoU results for PolypGen. Best results are \textbf{bold}.}
\label{six}
\resizebox{\textwidth}{!}{
\begin{tabular}{c c cc cc cc cc cc cc cc}
\toprule
\textbf{Model} & \textbf{Method} 
  & \multicolumn{2}{c}{\textbf{data\_C1}} 
  & \multicolumn{2}{c}{\textbf{data\_C2}} 
  & \multicolumn{2}{c}{\textbf{data\_C3}} 
  & \multicolumn{2}{c}{\textbf{data\_C4}} 
  & \multicolumn{2}{c}{\textbf{data\_C5}} 
  & \multicolumn{2}{c}{\textbf{data\_C6}} 
  & \textbf{mDice} & \textbf{mIoU} \\
 &  
  & \textbf{mDice} & \textbf{mIoU} 
  & \textbf{mDice} & \textbf{mIoU} 
  & \textbf{mDice} & \textbf{mIoU} 
  & \textbf{mDice} & \textbf{mIoU} 
  & \textbf{mDice} & \textbf{mIoU} 
  & \textbf{mDice} & \textbf{mIoU} 
  & \textbf{Overall} & \textbf{Overall} \\
\midrule
\multirow{6}{*}{PraNet} 
 & Baseline 
   & 79.92 & 73.35 
   & 73.74 & 68.15 
   & 87.44 & 80.95 
   & 36.50 & 31.65 
   & 56.96 & 48.63 
   & 75.92 & 70.24 
   & 71.20 & 64.91 \\
 & \textcolor{black}{Pixel to Pixel} 
   & \textcolor{black}{78.70*} & \textcolor{black}{71.40*} 
   & \textcolor{black}{72.79*} & \textcolor{black}{66.84*} 
   & \textcolor{black}{87.03*} & \textcolor{black}{80.12*} 
   & \textcolor{black}{36.85} & \textcolor{black}{31.65} 
   & \textcolor{black}{53.36*} & \textcolor{black}{45.73*} 
   & \textcolor{black}{73.43*} & \textcolor{black}{67.35*} 
   & \textcolor{black}{70.10} & \textcolor{black}{63.52} \\
 & \textcolor{black}{ArSDM} 
   & \textcolor{black}{81.13*} & \textcolor{black}{74.17} 
   & \textcolor{black}{72.84} & \textcolor{black}{67.24} 
   & \textcolor{black}{86.31} & \textcolor{black}{79.49*} 
   & \textcolor{black}{36.17} & \textcolor{black}{31.01*} 
   & \textcolor{black}{59.55*} & \textcolor{black}{50.97*} 
   & \textcolor{black}{74.78} & \textcolor{black}{68.40*} 
   & \textcolor{black}{71.12} & \textcolor{black}{64.55} \\
 & \textcolor{black}{Mask\_C} 
   & \textcolor{black}{79.47*} & \textcolor{black}{72.24*} 
   & \textcolor{black}{73.20*} & \textcolor{black}{67.27*} 
   & \textcolor{black}{86.87*} & \textcolor{black}{80.21*} 
   & \textcolor{black}{37.69*} & \textcolor{black}{32.22} 
   & \textcolor{black}{56.99} & \textcolor{black}{48.85} 
   & \textcolor{black}{\textbf{76.97}} & \textcolor{black}{\textbf{70.65}} 
   & \textcolor{black}{71.08} & \textcolor{black}{64.47} \\
   & $\text{PSDM}^{\text{S}}$ 
   & 80.75\textcolor{black}{*} & 73.92\textcolor{black}{*} 
   & \textbf{74.05}\textcolor{black}{*} & \textbf{68.31} 
   & \textbf{87.51}\textcolor{black}{*} & \textbf{81.06} 
   & 37.80\textcolor{black}{*} & 32.52\textcolor{black}{*} 
   & 58.03\textcolor{black}{*} & 50.08\textcolor{black}{*} 
   & 76.07\textcolor{black}{*} & 70.01 
   & 71.76 & 65.38 \\

   & $\text{PSDM}^{\text{M}}$ 
   & \textbf{81.89}\textcolor{black}{*} & \textbf{75.01} 
   & 73.60 & 67.92\textcolor{black}{*} 
   & 87.35 & 80.78 
   & \textbf{39.52}\textcolor{black}{*} & \textbf{34.26}\textcolor{black}{*} 
   & \textbf{60.80}\textcolor{black}{*} & \textbf{52.54}\textcolor{black}{*} 
   & 76.70 & 70.25 
   & \textbf{72.48} & \textbf{66.01} \\
\midrule
\multirow{6}{*}{Polyp-PVT} 
 & Baseline 
   & 82.58 & 74.37 
   & 74.17 & 68.09 
   & 87.48 & 80.21 
   & 40.78 & 34.46 
   & 63.90 & 53.82 
   & 77.11 & 70.81 
   & 73.38 & 66.00 \\
 & \textcolor{black}{Pixel to Pixel} 
   & \textcolor{black}{84.86*} & \textcolor{black}{77.25*} 
   & \textcolor{black}{75.30*} & \textcolor{black}{69.46} 
   & \textcolor{black}{88.35*} & \textcolor{black}{81.45} 
   & \textcolor{black}{41.76} & \textcolor{black}{35.56} 
   & \textcolor{black}{65.00} & \textcolor{black}{55.18*} 
   & \textcolor{black}{77.42*} & \textcolor{black}{71.02*} 
   & \textcolor{black}{74.55} & \textcolor{black}{67.47} \\
 & \textcolor{black}{ArSDM} 
   & \textcolor{black}{84.89} & \textcolor{black}{77.55} 
   & \textbf{\textcolor{black}{76.66}}\textcolor{black}{*} & \textbf{\textcolor{black}{71.20}} 
   & \textbf{\textcolor{black}{89.03}} & \textbf{\textcolor{black}{82.75}} 
   & \textcolor{black}{42.67} & \textcolor{black}{36.86} 
   & \textcolor{black}{66.02} & \textcolor{black}{57.03} 
   & \textbf{\textcolor{black}{79.46}}\textcolor{black}{*} & \textbf{\textcolor{black}{73.61}}\textcolor{black}{*} 
   & \textcolor{black}{75.40} & \textbf{\textcolor{black}{68.84}} \\
 & \textcolor{black}{Mask\_C} 
   & \textcolor{black}{84.89*} & \textcolor{black}{77.50*} 
   & \textcolor{black}{76.17*} & \textcolor{black}{70.57} 
   & \textcolor{black}{88.93*} & \textcolor{black}{82.40*} 
   & \textcolor{black}{43.20} & \textcolor{black}{37.12} 
   & \textcolor{black}{65.40} & \textcolor{black}{56.33} 
   & \textcolor{black}{77.76} & \textcolor{black}{71.12} 
   & \textcolor{black}{75.36} & \textcolor{black}{68.45} \\
   & $\text{PSDM}^{\text{S}}$ 
   & \textbf{85.11}\textcolor{black}{*} & 77.57\textcolor{black}{*} 
   & 75.84\textcolor{black}{*} & 69.97\textcolor{black}{*} 
   & 88.76\textcolor{black}{*} & 82.00\textcolor{black}{*} 
   & \textbf{44.28}\textcolor{black}{*} & \textbf{38.16}\textcolor{black}{*} 
   & 67.11\textcolor{black}{*} & 57.55\textcolor{black}{*} 
   & 76.56\textcolor{black}{*} & 70.64\textcolor{black}{*} 
   & 75.42 & 68.47 \\

   & $\text{PSDM}^{\text{M}}$ 
   & 84.77\textcolor{black}{*} & \textbf{77.60}\textcolor{black}{*} 
   & 76.23\textcolor{black}{*} & 70.69\textcolor{black}{*} 
   & 88.93\textcolor{black}{*} & 82.41\textcolor{black}{*} 
   & 43.86\textcolor{black}{*} & 37.73\textcolor{black}{*} 
   & \textbf{68.03}\textcolor{black}{*} & \textbf{58.71}\textcolor{black}{*} 
   & 76.69\textcolor{black}{*} & 70.80\textcolor{black}{*} 
   & \textbf{75.56} & \textbf{68.84} \\
\midrule
\multirow{6}{*}{Polyp-CASCADE} 
 & Baseline 
   & 85.72 & 78.79 
   & 75.78 & 70.38 
   & 89.92 & 83.86 
   & 44.10 & 38.62 
   & 64.44 & 55.86 
   & 77.87 & 72.12 
   & 75.55 & 69.23 \\
 & \textcolor{black}{Pixel to Pixel} 
   & \textcolor{black}{85.24*} & \textcolor{black}{78.63*} 
   & \textcolor{black}{\textbf{77.04}} & \textcolor{black}{\textbf{71.70}} 
   & \textcolor{black}{90.06*} & \textcolor{black}{84.02*} 
   & \textcolor{black}{44.45} & \textcolor{black}{38.25} 
   & \textcolor{black}{64.24*} & \textcolor{black}{55.30} 
   & \textcolor{black}{78.25*} & \textcolor{black}{72.98*} 
   & \textcolor{black}{75.82} & \textcolor{black}{69.43} \\
 & \textcolor{black}{ArSDM} 
   & \textcolor{black}{85.97*} & \textcolor{black}{79.19*} 
   & \textcolor{black}{76.32} & \textcolor{black}{71.00} 
   & \textcolor{black}{90.01} & \textcolor{black}{83.88*} 
   & \textcolor{black}{43.81} & \textcolor{black}{38.06*} 
   & \textcolor{black}{66.06*} & \textcolor{black}{57.19*} 
   & \textbf{\textcolor{black}{80.52}}\textcolor{black}{*} & \textbf{\textcolor{black}{74.73}}\textcolor{black}{*}
   & \textcolor{black}{76.04} & \textcolor{black}{69.67} \\
 & \textcolor{black}{Mask\_C} 
   & \textcolor{black}{86.08*} & \textcolor{black}{79.36*} 
   & \textcolor{black}{76.48*} & \textcolor{black}{71.32*} 
   & \textcolor{black}{89.82*} & \textcolor{black}{83.76} 
   & \textcolor{black}{43.15*} & \textcolor{black}{37.43*} 
   & \textcolor{black}{63.61*} & \textcolor{black}{54.95*} 
   & \textcolor{black}{78.79} & \textcolor{black}{73.21} 
   & \textcolor{black}{75.51} & \textcolor{black}{69.24} \\
 & $\text{PSDM}^{\text{S}}$ 
   & 85.77\textcolor{black}{*} & 79.12\textcolor{black}{*} 
   & 76.17\textcolor{black}{*} & 70.85\textcolor{black}{*} 
   & \textbf{90.36}\textcolor{black}{*} & \textbf{84.37}\textcolor{black}{*} 
   & 44.35 & 38.67\textcolor{black}{*} 
   & 67.13\textcolor{black}{*} & 58.59\textcolor{black}{*} 
   & 78.23\textcolor{black}{*} & 72.65\textcolor{black}{*} 
   & 76.18 & 69.94 \\

 & $\text{PSDM}^{\text{M}}$ 
   & \textbf{86.17}\textcolor{black}{*} & \textbf{79.68}\textcolor{black}{*} 
   & 76.37 & 71.11 
   & 89.71 & 83.64 
   & \textbf{45.25}\textcolor{black}{*} & \textbf{39.94}\textcolor{black}{*} 
   & \textbf{68.04}\textcolor{black}{*} & \textbf{58.88}\textcolor{black}{*} 
   & 77.67\textcolor{black}{*} & 72.02 
   & \textbf{76.32} & \textbf{70.06} \\
\bottomrule
\end{tabular}
}
\end{table*}

\subsection{Experiments Results}

\subsubsection{Polyp Segmentation}

In our experiments, we evaluated the performance of the segmentation models using two key metrics: mean Dice coefficient (mDice) and mean Intersection over Union (mIoU). These metrics are crucial in medical image segmentation, as they measure the overlap between the predicted segmentations and the ground truth, with higher values indicating better segmentation performance. \textcolor{black}{Statistical significance was evaluated using two-sided Wilcoxon signed-rank tests, with significant differences (p$<$0.05) indicated by an asterisk (*) in the Table~\ref{five} and Table~\ref{six}.}

\textbf{Impact on Generalization:} 
The results indicate a consistent improvement in both mDice and mIoU scores across models and datasets with each augmentation strategy, underscoring the effectiveness of our approach in enhancing model generalization. It is important to note that the $\text{PSDM}^\text{S}$ strategy refers to pretraining the models using the \textit{Segmentation Mask Only} set, where only mask annotations are employed, focusing purely on spatial structure. In contrast, the $\text{PSDM}^\text{M}$ strategy involves pretraining the models using a combination of both the \textit{Segmentation Mask Only} set and the \textit{Segmentation Mask-Text Description} set. Specifically, across the CVC-300, Clinic-DB, Kvasir, CVC-ColonDB, and ETIS datasets, the best performance was generally observed when using the  $\text{PSDM}^\text{M}$ augmentation strategy. This approach, which combines both segmentation masks and textual descriptions through compositional prompts, significantly aids in capturing finer details and enhancing spatial contextualization, leading to better segmentation outcomes. For instance, PraNet's overall mDice increased from 74.27 (Baseline) to 76.70 ($\text{PSDM}^\text{M}$), while Polyp-PVT and Polyp-CASCADE showed similar trends, with improvements in both mDice and mIoU across the board. These results further highlight the utility of augmentation strategies, particularly the  $\text{PSDM}^\text{M}$  approach, in improving the generalization capabilities of the models, enabling them to segment polyps more accurately (see Table \ref{five}).

\textbf{Evaluation on Complex Datasets:} 
The models demonstrated robust performance on the PolypGen test set, as detailed in Table \ref{six}. It's important to note the distinction between the experiments presented in Tables \ref{five} and \ref{six}. Table \ref{five} focuses on evaluating the models across standard segmentation datasets like CVC-300 and Clinic-DB, which are commonly used benchmarks in polyp segmentation research. In contrast, Table \ref{six} presents the results on the PolypGen dataset, which includes more challenging subsets (dataC4 and dataC5) known for their diverse polyp appearances. Notably, the proposed method showed substantial improvements on more challenging datasets such as ETIS and the PolypGen subsets dataC4 and dataC5. Polyp-CASCADE, in particular, achieved one of the most significant performance boosts on the PolypGen dataC5 subset, reflecting its enhanced ability to generalize to cases with smaller or irregularly shaped polyps.

Overall, these findings suggest that the proposed augmentation strategies consistently enhance models' ability to generalize across both standard and challenging datasets. \textcolor{black}{Compared with augmentations using polyp images from alternative generative models, our approach uniformly outperforms those augmentations across all evaluated segmentation architectures, yielding the highest overall mDice and mIoU on both standard benchmarks and the out‐of‐distribution PolypGen dataset.} By integrating spatial and textual information through compositional prompts, the models show a superior capacity to handle variability in polyp characteristics. This improved generalization is crucial for real-world medical applications, where polyps exhibit significant heterogeneity in appearance, and accurate segmentation plays  vital role in diagnostic precision.

\subsubsection{Polyp Classification}

The dataset used for the classification task consisted of 308 benign and 61 malignant polyp images, presenting a significant class imbalance with malignant polyps being underrepresented. This imbalance posed challenges for model training, particularly in detecsting malignant polyps, which are critical for clinical decision-making.

\begin{figure}[ht]
    \centering
    \includegraphics[width=0.8\linewidth]{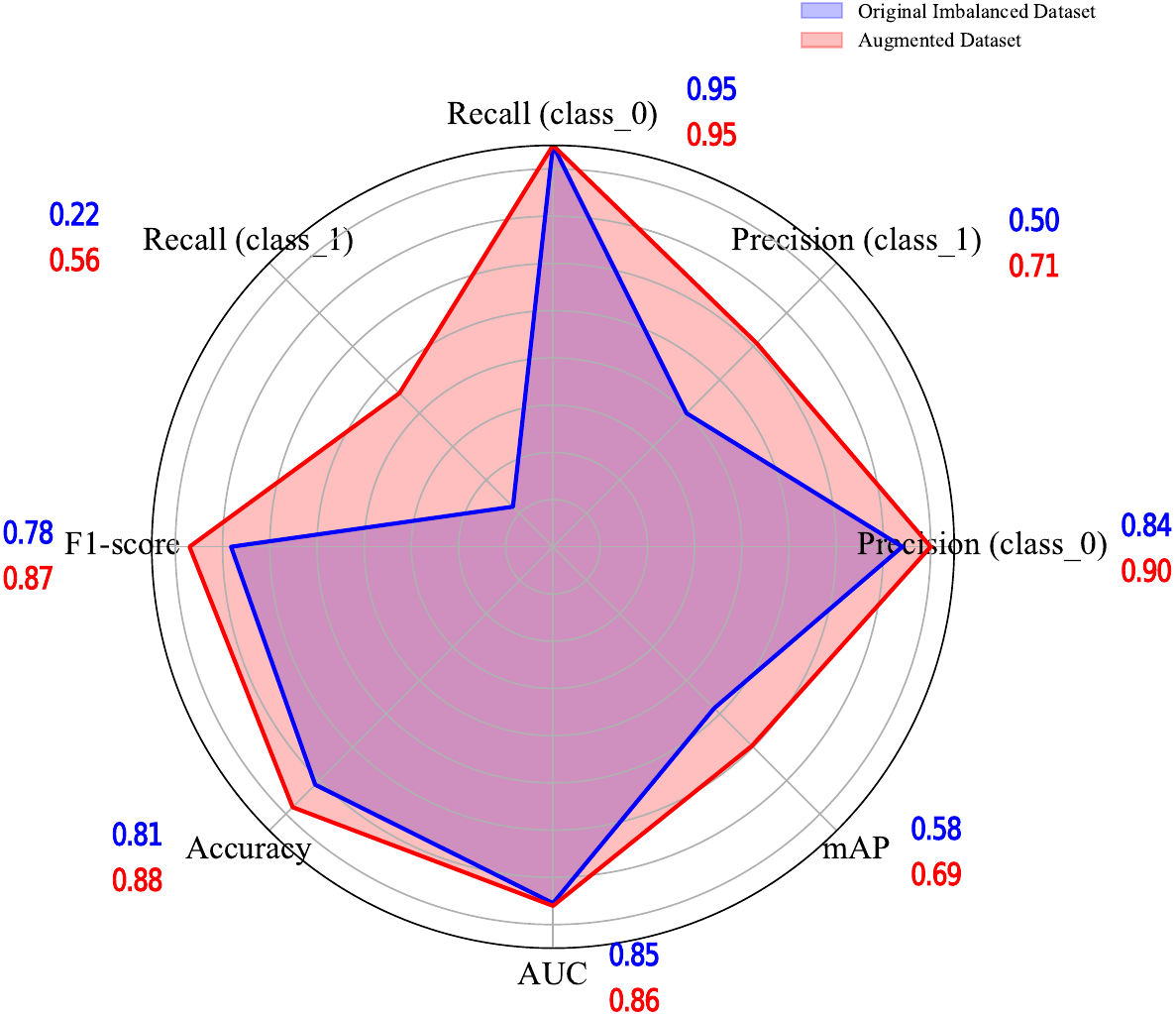}
    \caption{\textcolor{black}{Radar chart illustrating the performance comparison between ResNet models trained on the original imbalanced dataset and the augmented balanced dataset, with exact metric values overlaid.}}
    \label{resnet}
\end{figure}
\begin{figure}[ht]
    \centering
    \includegraphics[width=1.0\linewidth]{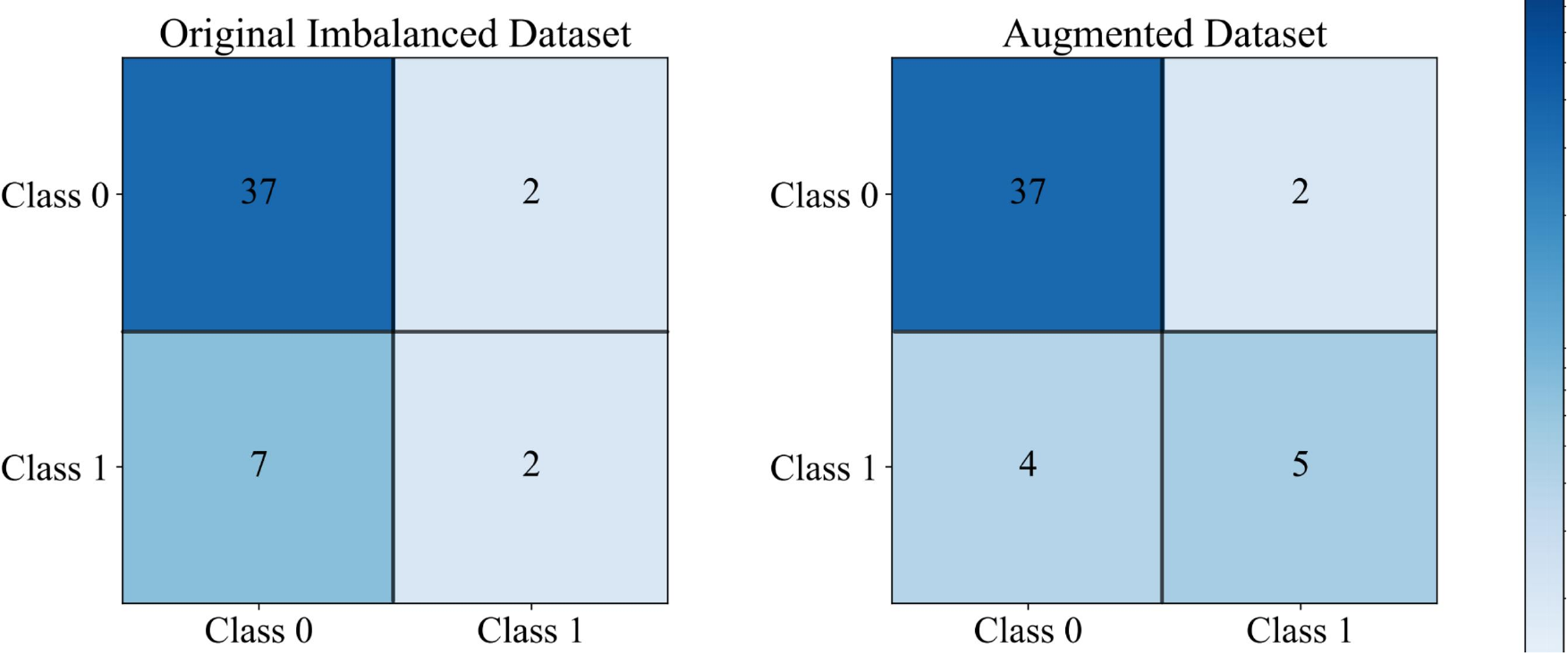}
    \caption{Comparison of Confusion Matrices for Classification Performance the original and augmented dataset performance.}
    \label{resnet_metric}
\end{figure}

\begin{figure*}[ht]
    \centering
    \includegraphics[width=1.0\linewidth]{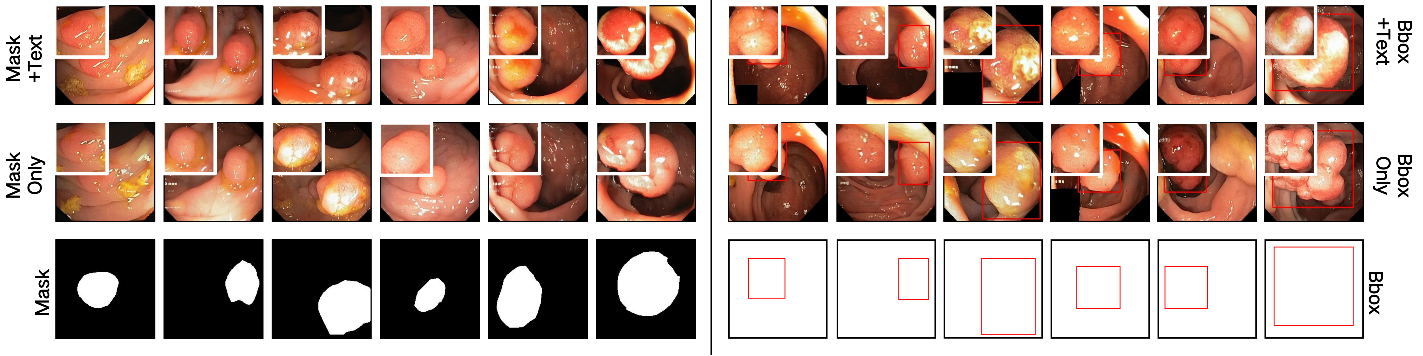}
    \caption{\textcolor{black}{Left: Images generated with masks alone and masks with a text description (``polyp, reddish color"). Right: Images generated with bounding boxes (Bbox) alone and Bbox with text descriptions (``polyp, the surface is congested and granular"). The top-left inset highlights the lesion region.
}}

    \label{fig:case_study_1}
\end{figure*}

In the original imbalanced dataset experiment, we evaluated the performance of a ResNet classification model trained on this imbalanced dataset. The class imbalance negatively impacted the model’s performance, resulting in lower accuracy, especially in identifying malignant polyps. \textcolor{black}{Moreover, simply replicating minority-class samples can lead to overfitting on} \textcolor{black}{duplicated instances, further degrading generalization~\cite{satpathy2023smote}.}
 To address this issue, we conducted an experiment using an augmented balanced dataset. Additional images of malignant polyps were generated using the prompt ``a V* polyp" creating a more balanced class distribution and mitigating the adverse effects of the original imbalance.

\begin{figure*}[htbp]
    \centering
    \includegraphics[width=0.95\linewidth]{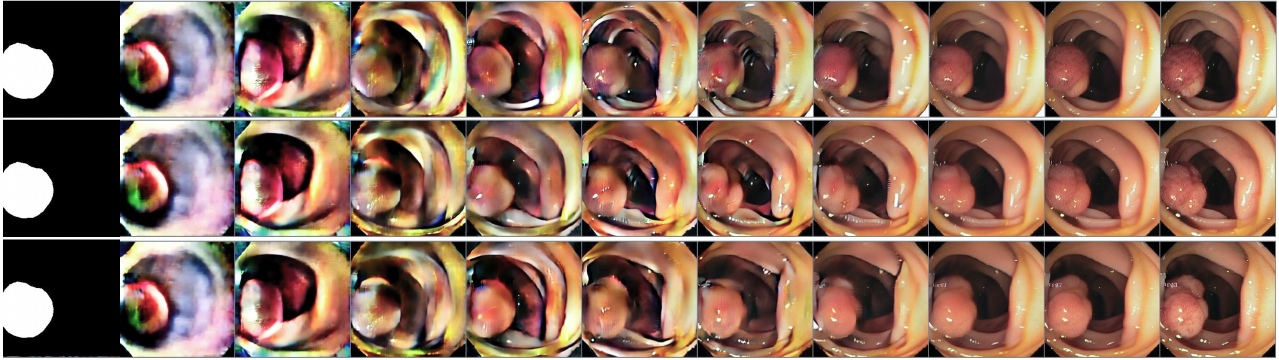}
    \caption{Each row in this figure demonstrates the generation process from the same segmentation mask using three different text descriptions: ``Polyp, 0-lsp," ``Polyp, with a lobed surface, local depression," and ``Polyp, with a short pedicle." }
    \label{fig:case_study_2}
\end{figure*}

The radar chart in Fig. \ref{resnet} and the confusion matrices in Fig. \ref{resnet_metric} illustrate the performance improvements of the ResNet model trained on the augmented balanced dataset compared to the original imbalanced dataset, showing enhanced accuracy and reduced false negatives for malignant polyps.
The results demonstrated that the ResNet model trained on the augmented balanced dataset exhibited substantial improvements, particularly in its ability to detect malignant polyps. The model's accuracy and sensitivity in identifying this minority class increased significantly, showcasing the effectiveness of balancing the dataset through augmentation. The balanced dataset also led to more consistent performance across various evaluation metrics, indicating improved generalization and robustness of the model. Overall, the ResNet model trained on this balanced dataset demonstrated superior accuracy and robustness, particularly in detecting malignant polyps.

\subsubsection{Polyp Detection}

\begin{table}[ht]
\centering
\caption{Performance Comparison of YOLOv5 on PolypGen}
\resizebox{0.40\textwidth}{!}{
\begin{tabular}{lccc}
\toprule
\multirow{2}{*}{\textbf{Method}} & \multicolumn{3}{c}{\textbf{PolypGen}} \\
\cmidrule(lr){2-4}
 & \textbf{F1 (\%)} & \textbf{mAP50 (\%)} & \textbf{mAP50-95 (\%)} \\
\midrule
Baseline & 71.79 & 73.34 & 57.74 \\
$\text{PSDM}^\text{B}$    & 71.70 & 74.06 & 59.10 \\
$\text{PSDM}^\text{M}$     & \textbf{73.91} & \textbf{76.24} & \textbf{60.83} \\
\bottomrule
\end{tabular}
}
\label{polyp_detection}
\end{table}

\begin{table}[ht]
\centering
\caption{\textcolor{black}{Quantitative Evaluation Metrics}}
\resizebox{0.40\textwidth}{!}{%
  \begin{tabular}{@{}%
      >{\centering\arraybackslash}p{3.0cm}%
      @{\hskip 0.5cm}%
      >{\centering\arraybackslash}p{2.5cm}%
      @{\hskip 0.5cm}%
      >{\centering\arraybackslash}p{1.5cm}%
    @{}}
    \toprule
    \multirow{2}{*}{\textbf{Method}} & \multicolumn{2}{c}{\textbf{Metrics}} \\
    \cmidrule(lr){2-3}
     & \textbf{FID ↓} & \textbf{SSIM ↑} \\
    \midrule
    ArSDM            & 200.35 & 0.4489 \\
    Mask\_C          & 175.93 & \textbf{0.6115} \\
    Pix2pix          & 271.44 & 0.5759 \\
    $\text{PSDM}^\text{S}$ (Ours) & \underline{120.64} & 0.5770 \\
    $\text{PSDM}^\text{M}$ (Ours) & \textbf{111.52} & \underline{0.5839} \\
    \bottomrule
  \end{tabular}
}
\label{tab:FID}
\end{table}

In addition to polyp segmentation and classification experiments, we conducted a polyp detection experiment using the YOLOv5 model to further validate the effectiveness of our compositional prompt-guided generation approach. For this experiment, we utilized two augmented datasets—the \textit{Bounding Box Only} set and the \textit{Bounding Box-Text Description} set.
The results, summarized in Table \ref{polyp_detection}, indicate consistent improvements with the compositional prompt-guided augmentation. Specifically, the  $\text{PSDM}^\text{M}$ strategy yielded the highest F1 score (73.91\%) and mAP50-95 (60.83\%) on the PolypGen dataset, surpassing both the Baseline and $\text{PSDM}^\text{B}$ augmentation strategies. This enhancement demonstrates that integrating both bounding boxes and textual descriptions in the augmentation process significantly boosts the model's ability to detect polyps accurately.

Moreover, the improvements in mAP50 and mAP50-95 metrics, as shown in Table \ref{polyp_detection}, reflect the model's enhanced ability to detect polyps across different IoU thresholds. A higher mAP50 indicates better performance at the standard IoU threshold of 0.5, while the increased mAP50-95 suggests robustness over a range of stricter IoU criteria. This robustness is essential for detecting polyps of varying sizes and shapes, which are common challenges in real-world clinical settings.

\subsubsection{Qualitative Diversity Evaluation}
Fig. \ref{fig:case_study_1} visually evaluates the diversity of generated samples, showcasing distinct textures, shapes, and sizes of polyps created using compositional prompts. \textcolor{black}{The enlarged lesion regions highlight that by introducing semantic control through textual prompts, we can generate more diverse samples.} Fig. \ref{fig:case_study_2} illustrates the transition during denoising, where the segmentation mask initially guides low-frequency structures, such as shape and position, while text descriptions refine high-frequency details like texture and pathology. This transition is controlled by a weighting scheme, with the text prompt weight \( \lambda_t^{\text{text}} = (1 + \cos( \pi \cdot t / T )) / 2 \) increasing over time, and the mask prompt weight \( \lambda_t^{\text{mask}} = 1 - \lambda_t^{\text{text}} \) ensuring a smooth shift between prompts.

\subsubsection{Quality Assessment of Generated Samples}

\textcolor{black}{Table \ref{tab:FID} compares the quality of generated samples using FID and SSIM metrics. Our $\text{PSDM}^\text{M}$ achieved the lowest FID score (111.52), clearly outperforming other methods such as ArSDM (200.35) and Pix2pix (271.44), and moderately better than our own $\text{PSDM}^\text{S}$ (120.64). In terms of SSIM, $\text{PSDM}^\text{M}$ (0.5839) was second-best, close to Mask\_C (0.6115), and notably} \textcolor{black}{ higher than ArSDM (0.4489). These findings confirm that our approach provides good balance between visual fidelity and structural accuracy.}

\textcolor{black}{To further assess the clinical realism of our PSDM-generated images, we performed a blinded evaluation in which} \textcolor{black}{an experienced gastroenterologist (15 years of endoscopy practice) rated each image’s realism on a 5-point Likert scale. We randomly selected 150 images from each of two sets: \textit{Segmentation Mask Only} set ($\text{PSDM}^{\text{S}}$) and \textit{Segmentation Mask-Text Description} set ($\text{PSDM}^{\text{M}}$). Both sets yielded mean realism scores above 4, with $\text{PSDM}^{\text{S}}$ achieving a mean of 4.347 and $\text{PSDM}^{\text{M}}$ a mean of 4.447.}

{\color{black}
\subsection{Ablation Studies on Prompt Integration and Weighting Schemes}

To validate our coarse‐to‐fine compositional prompt strategy and assess the influence of different $\lambda$‐schedules (Eq.~\ref{lambda}), we conducted two ablation studies: (i) measuring the efficacy of transitioning from coarse to fine prompts, and (ii) quantifying the impact of $\lambda$‐schedules on LPIPS and 1-SSIM. Because prompts at the same granularity are alternatives—segmentation masks and bounding boxes both encode spatial layout, and prototype images are needed only when textual descriptions are unavailable. We therefore examine two non‐overlapping combinations (\emph{mask + text} and \emph{bbox + vague class}) to highlight how integrating coarse localization with fine semantic cues enhances generation quality.

\subsubsection{Effectiveness of Coarse-to-Fine Prompt Integration}

We examine two coarse-to-fine transitions—segmentation mask $\rightarrow$ text prompts and bounding box $\rightarrow$ vague class prompts. Table~\ref{tab:ablation_seg_text} compares text-only, mask-only, and several mask $\rightarrow$ text schedules. Text-only guidance yields the highest semantic diversity (LPIPS = 0.6294) but the lowest structural coherence (1–SSIM = 0.8836), whereas mask-only guidance ensures structural fidelity (1–SSIM = 0.8867) at the expense of diversity (LPIPS = 0.5815). Among the mixed schemes, the linear (mask $\rightarrow$ text) schedule—which linearly decreases mask weight from 1 to 0 while increasing text weight from 0 to 1 over 200 diffusion steps—best balances both metrics (LPIPS = 0.5953, 1–SSIM = 0.9260). In contrast, the reverse linear (text $\rightarrow$ mask) transition underperforms (LPIPS = 0.5893, 1–SSIM = 0.8920) because later mask enforcement overrides earlier semantic cues.
Likewise, Table~\ref{tab:ablation_bbox_vague} shows that vague class-only prompts produce poor diversity and coherence (LPIPS = 0.3887, 1–SSIM = 0.6085), while bbox-only prompts secure strong structure (1–SSIM = 0.8823) but lack semantic richness (LPIPS = 0.5914). All combined bbox $\rightarrow$ vague class schedules outperform single-cue baselines, with linear (bbox $\rightarrow$ vague class) achieving the best trade-off (LPIPS = 0.6294, 1–SSIM = 0.8866) by preserving spatial layout initially and gradually introducing semantic detail.
These results rigorously confirm that a coarse-to-fine progression—first applying structural guidance (mask or bbox) then semantic refinement (text or vague class)—optimally enhances both the structural coherence and semantic diversity of generated polyp images.}

\begin{table}[htbp]
\centering
\caption{\textcolor{black}{Impact of segmentation mask and textual prompts on image generation. Best results are \textbf{bold}, second-best are \underline{underlined}.}}
\label{tab:ablation_seg_text}
\resizebox{0.48\textwidth}{!}{%
\begin{tabular}{lccccc}
\toprule
  & Text & Mask & $\lambda$ Scheme & LPIPS ($\uparrow$) & 1--SSIM ($\uparrow$) \\
\midrule
\multirow{2}{*}{Single}        
    & \checkmark & –           & –                  & \textbf{0.6294±0.0687} & 0.8836±0.1090 \\
    & –          & \checkmark  & –                  & 0.5815±0.0743         & 0.8867±0.1135 \\
\midrule
\multirow{4}{*}{Compositional}
    & \checkmark & \checkmark  & stage (mask→text)  & 0.5922±0.0708         & 0.8884±0.1150 \\
    & \checkmark & \checkmark  & cosine (mask→text) & 0.5947±0.0645         & \underline{0.8969±0.1054} \\
    & \checkmark & \checkmark  & linear (mask→text) & \underline{0.5953±0.0655} & \textbf{0.9260±0.0860} \\
    & \checkmark & \checkmark  & linear (text→mask)& 0.5893±0.0679         & 0.8920±0.1121 \\
\bottomrule
\end{tabular}}%
\end{table}

\begin{table}[htbp]
\centering
\caption{\textcolor{black}{Impact of bounding box and vague class prompts on image generation. Best results are \textbf{bold}, second-best are \underline{underlined}.}}
\label{tab:ablation_bbox_vague}
\resizebox{0.48\textwidth}{!}{%
\begin{tabular}{lccccc}
\toprule
  & Vague Class & Bbox & $\lambda$ Scheme & LPIPS ($\uparrow$) & 1--SSIM ($\uparrow$) \\
\midrule
\multirow{2}{*}{Single}
    & \checkmark & –           & –                         & 0.3887±0.0280         & 0.6085±0.0611 \\
    & –          & \checkmark  & –                         & 0.5914±0.0833         & 0.8823±0.0843 \\
\midrule
\multirow{4}{*}{Compositional}
    & \checkmark & \checkmark  & stage (bbox→class)        & 0.6037±0.0739         & 0.8827±0.0983 \\
    & \checkmark & \checkmark  & cosine (bbox→class)       & \underline{0.6146±0.0469} & \underline{0.8857±0.0644} \\
    & \checkmark & \checkmark  & linear (bbox→class)       & \textbf{0.6294±0.0687} & \textbf{0.8866±0.1090} \\
    & \checkmark & \checkmark  & linear (class→bbox)       & 0.6013±0.0366         & 0.8822±0.0525 \\
\bottomrule
\end{tabular}}%
\end{table}

\subsubsection{Impact of Weighting Hyperparameters}

\textcolor{black}{In Eq.~\eqref{lambda}, we specify four \(\lambda\)-schedules for guiding the transition between coarse and fine prompts. The \emph{stage (mask\(\rightarrow\)text)} schedule maintains full mask guidance for diffusion steps 1–100 before switching entirely to text guidance for steps 101–200; the \emph{cosine (mask\(\rightarrow\)text)} schedule smoothly interpolates mask weight downward and text weight upward following a cosine curve over all 200 steps; the \emph{linear (mask\(\rightarrow\)text)} schedule decreases mask weight linearly from 1 to 0 while increasing text weight from 0 to 1 throughout the diffusion process; and the \emph{linear (text\(\rightarrow\)mask)} schedule performs the exact inverse linear interpolation. Analogous definitions apply to bounding box and vague class prompts, yielding \emph{stage (bbox\(\rightarrow\)class)}, \emph{cosine (bbox\(\rightarrow\)class)}, \emph{linear (bbox\(\rightarrow\)class)}, and \emph{linear (class\(\rightarrow\)bbox)}. As shown in Tables~\ref{tab:ablation_seg_text} and \ref{tab:ablation_bbox_vague}, the linear coarse-to-fine schedules (mask\(\rightarrow\)text and bbox\(\rightarrow\)class) consistently achieve superior LPIPS and 1–SSIM, outperforming both the stage and cosine transitions as well as any fine-to-coarse reversal.
}

\begin{figure}
    \centering
    \includegraphics[width=1.0\linewidth]{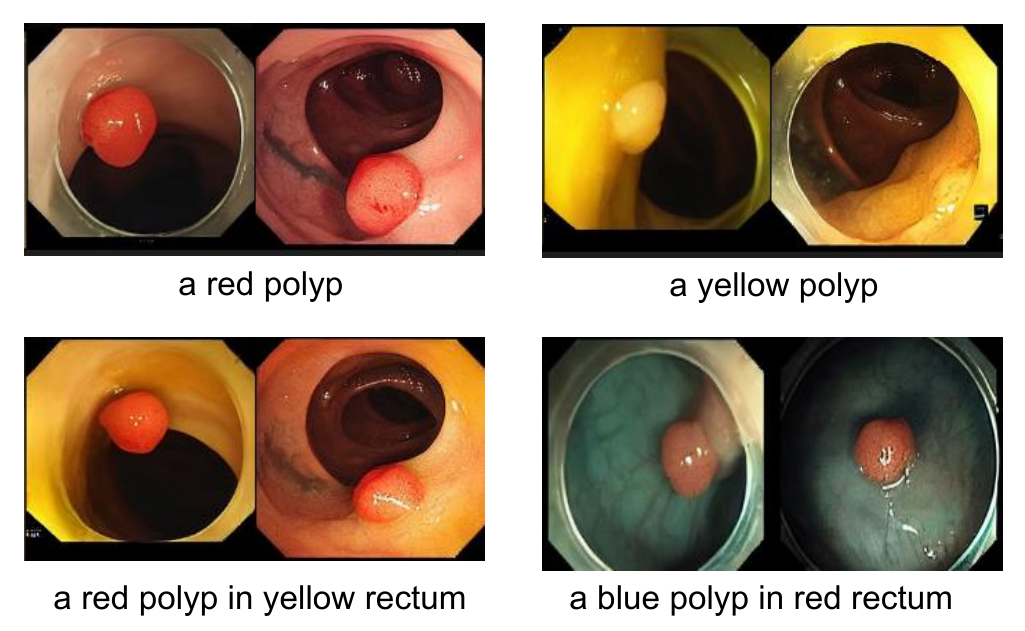}
    \caption{\textcolor{black}{Examples of unsuccessful synthetic polyp images generated by our model.}}
    \label{fig:failure_samples}
\end{figure}

\section{Discussion}
Recent advancements in generative models have shown promise for natural image synthesis \cite{dhariwal2021diffusion}. However, challenges remain in medical imaging, particularly due to the complexity and variability of medical data \cite{boehm2022harnessing}. Medical data are dynamic, evolving over time and varying across institutions, limiting the availability of comprehensive longitudinal datasets for training robust models. To address these challenges, our $\text{PSDM}^\text{M}$ strategy integrates multimodal data through compositional prompts, demonstrating substantial benefits, especially in complex datasets like PolypGen.
\textcolor{black}{The robustness of our approach is further evidenced by cross-LLM comparisons. Specifically, when substituting GPT-4o with DeepSeek-V2 to extract key attributes from text using a structured, template-based prompting strategy, both models consistently achieved error-free performance.}
\textcolor{black}{Our PSDM approach also demonstrates broad applicability beyond polyp detection and diagnosis, as it imposes no restrictions on image modality—such as CT, MRI, histopathology, or endoscopy—provided multi-modal priors (e.g., masks, boxes, reports) are available.}

\textcolor{black}{A limitation of our approach is a slight decrease in training‐set performance—for example, the Polyp‐CASCADE model’s mDice on Kvasir declines from 93.09\% to 92.01\%, as shown in Table~\ref{five}.This decrease highlights that, although our synthetic augmentation markedly strengthens robustness to unseen PolypGen samples, it can slightly impair performance on data drawn from the original training distribution. The baseline model retains higher in‐distribution accuracy by relying solely on familiar examples but fails to generalize to clinically diverse scenarios.}
\textcolor{black}{Besides, we illustrate some failure samples of our method in Fig.~\ref{fig:failure_samples}, which demonstrate that underrepresented or weakly supported prompts (e.g., “yellow polyp” or “blue polyp”) often lead to indiscriminate colorization of the entire scene or outright generation failure, reflecting a weak prior for rare color–lesion combinations and an inability to disentangle the foreground lesion from the mucosal background when co‐occurrence statistics are scarce; notably, more explicit, compound prompts such as “red polyp in yellow rectum” provide stronger conditional cues that partially restore correct localization and contrast.  }

Looking forward, integrating synthetic data into clinical workflows presents a promising yet challenging frontier. Synthetic datasets offer opportunities to address class imbalances and simulate rare conditions. However, ensuring clinical interpretability, reliability, and utility for decision-making will be crucial for adoption in healthcare settings. Future research should focus on robust evaluation frameworks to assess the visual quality and clinical relevance of synthetic images, \textcolor{black}{as well as dynamic strategies to balance domain-specific performance with cross-institutional generalization}.

\section{Conclusion}
In this work, we introduced a novel compositional prompt-guided diffusion model designed to integrate independently annotated datasets for colorectal cancer (CRC) imaging. By leveraging our Progressive Spectrum Diffusion Model (PSDM), which utilizes a frequency-based prompt spectrum to progressively refine images from coarse spatial structures to fine-grained details, we successfully enhanced the model’s performance by generating more diverse polyp images through the incorporation of richer clinical information. The resulting images significantly improved polyp segmentation, detection, and classification tasks, ultimately contributing to more accurate diagnosis of colorectal cancer.



\end{document}